\documentclass[10pt,twocolumn,letterpaper]{article}

\usepackage[pagenumbers]{cvpr} 

\definecolor{cvprblue}{rgb}{0.21,0.49,0.74}
\usepackage[pagebackref,breaklinks,colorlinks,allcolors=cvprblue]{hyperref}
\usepackage{algorithm}
\usepackage{algorithmic}
\usepackage{multirow}
\usepackage{wrapfig}

\title{Use as Many Surrogates as You Want: Selective Ensemble Attack to Unleash Transferability without Sacrificing Resource Efficiency}

\author{%
	Bo Yang\textsuperscript{1\thanks{Equal contribution.}}, 
        Hengwei Zhang\textsuperscript{1\textsuperscript{$\ast$$\dagger$}}, 
        Jindong Wang\textsuperscript{1}, 
        Yuchen Ren\textsuperscript{2}, \\
        Chenhao Lin\textsuperscript{2},
        Chao Shen\textsuperscript{2},
        Zhengyu Zhao\textsuperscript{2\thanks{Corresponding author.}}\\
	\textsuperscript{1}Information Engineering University, China;
	\textsuperscript{2}Xi’an Jiaotong University, China \\
	{\tt\small \{yangbo\_hn, wlby\_zzmy\_henan\}@163.com}, 
    {\tt\small zhengyu.zhao@xjtu.edu.cn}
}

\begin{document}
\maketitle

\begin{abstract}

In surrogate ensemble attacks, using more surrogate models yields higher transferability but lower resource efficiency.
This practical trade-off between transferability and efficiency has largely limited existing attacks despite many pre-trained models are easily accessible online.
In this paper, we argue that such a trade-off is caused by an unnecessary common assumption, i.e., all models should be \textit{identical} across iterations.
By lifting this assumption, we can use as many surrogates as we want to unleash transferability without sacrificing efficiency.
Concretely, we propose Selective Ensemble Attack (SEA), which dynamically selects \textit{diverse} models (from easily accessible pre-trained models) across iterations based on our new interpretation of decoupling within-iteration and cross-iteration model diversity.
In this way, the number of within-iteration models is fixed for maintaining efficiency, while only cross-iteration model diversity is increased for higher transferability.
Experiments on ImageNet demonstrate the superiority of SEA in various scenarios.
For example, when dynamically selecting 4 from 20 accessible models, SEA yields 8.5\% higher transferability than existing attacks under the same efficiency.
The superiority of SEA also generalizes to real-world systems, such as commercial vision APIs and large vision-language models. 
Overall, SEA opens up the possibility of adaptively balancing transferability and efficiency according to specific resource requirements.

\end{abstract}

\section{Introduction}
\label{sec:intro}

Deep Neural Networks (DNNs) have been shown to be strikingly vulnerable to adversarial examples, i.e., maliciously crafted samples by adding small perturbations to benign inputs~\cite{Szegedy2014Intriguing,Goodfellow2015Explaining}.
An intriguing property of adversarial examples that makes them threatening in the real world is their transferability~\cite{Carlini2017Towards,li2023towards,yang2025AES,Qin2022RAP,Andrew2019bugs,Zhang2022NAA}, meaning that samples crafted on (known) surrogate models can also mislead (unknown) target models.
Transferable attacks have been extensively studied, based on different strategies, such as gradient optimization, input transformation, and surrogate ensemble~\cite{zhao2025revisiting,gu2024survey}.

\begin{figure}[t]
  \centering
    \includegraphics[width=\linewidth]{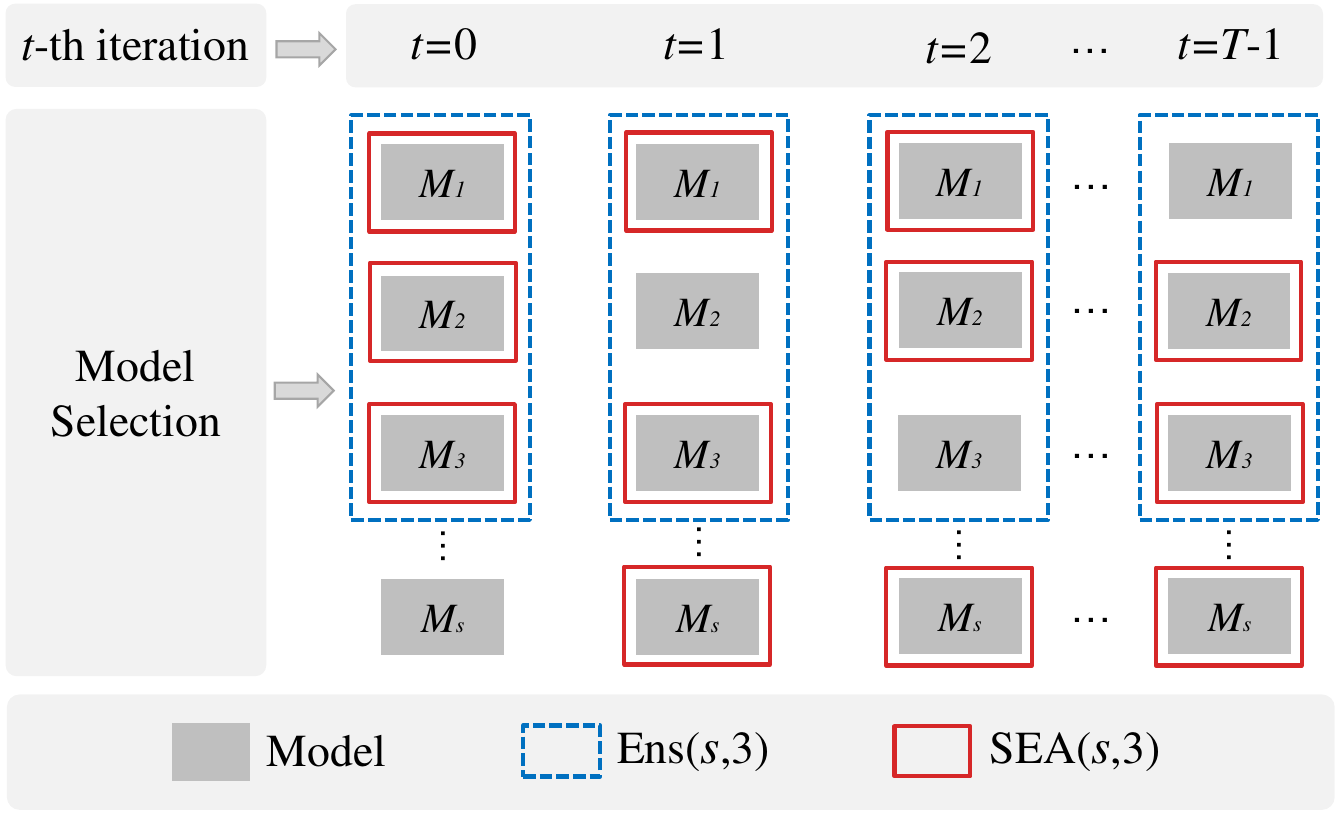}
        \caption{Our Selective Ensemble Attack (SEA) vs. conventional ensemble attacks (Ens) given $s$ easily accessible pre-trained models but restricted resources allowing only $m$ models per iteration. Our SEA($s$,$m$) dynamically selects $m$ \textcolor[HTML]{DC2728}{\textit{diverse}} models across iterations compared to \textcolor[HTML]{0070C3}{\textit{identical}} models in Ens($s$,$m$), leading to higher transferability. Here $m=3$ is used for illustration.}
    \label{fig:idea}
\end{figure}

Among them, surrogate ensemble is widely applied due to its simplicity and easy integration with other transfer techniques~\cite{Wang2021FIA,wu2020skip,zhu2023GRA,zhu2022DRA}.
In particular, there exists a well-known trade-off between attack transferability and the number of surrogate models, since using more models means lower resource efficiency regarding computational time and memory.
Resource efficiency is even more critical when using transferable adversarial attacks for adversarial training-based defenses~\cite{Florian2018EnsembleAT,yang2020dverge,sitawarinpubdef}.

To avoid high resource overhead, the common practice is to simply restrict the number of models to only a few, typically $<5$~\cite{Dong2018momentum,Lin2020Nesterov,long2022Frequency}.
Under such a restricted condition, although complex, optimized ensemble strategies~\cite{Xiong2021SVRE,chen2023AdaEA,tang2024ensemble,chen2024ensemble} can somehow help, the resulting transferability is still far from using many more models, according to our experimental results in Section~\ref{sec:Ens(20)} and a concurrent study~\cite{liu2024scaling} on scaling law of transferability.
In practice, it is meaningless to restrict the number of models since various pre-trained models are easily accessible online.

In this paper, rather than proposing a new ensemble strategy, we argue that the trade-off between transferability and resource efficiency can be removed, i.e., we can adopt as many surrogate models as we want to improve transferability without sacrificing resource efficiency.
This is made possible by our new interpretation of model diversity: decoupling model diversity into within-iteration and cross-iteration dimensions.
In general, both dimensions account for transferability, but only within-iteration diversity (corresponding to the number of models used in each iteration) accounts for resource efficiency.
Specifically, we quantify these two dimensions of diversity based on gradient similarity~\cite{kariyappa2019improving,li2020towards} to explore how they respectively contribute to the final transferability.

Based on the above new interpretation, we propose Selective Ensemble Attack (SEA), which dynamically selects a fixed number of \textit{diverse} models (from easily accessible pre-trained models) across iterations.
In this way, SEA increases cross-iteration diversity (by engaging more models across iterations) for higher transferability while maintaining within-iteration diversity (by fixing the number of models in each iteration).
In contrast, the common practice, i.e., \textit{identical} models should be adopted across iterations, blindly entangles the two dimensions of model diversity, and so it has to simultaneously increase transferability and resource costs.
Figure~\ref{fig:idea} illustrates the difference between our SEA and conventional ensemble.

In sum, the main contributions of this work are: 

\begin{itemize}[left=0pt]
\item We argue that the trade-off between transferability and resource efficiency in model ensemble attacks can be removed. We attribute the problem to entangled within- and cross-iteration model diversity, caused by the common practice that forces \textit{identical} models across iterations.

\item We propose Selective Ensemble Attack (SEA), which dynamically selects \textit{diverse} models across iterations to leverage as many models as we want to unleash transferability without sacrificing resource efficiency.
    
\item Extensive experiments on ImageNet validate the superiority of our SEA in both transferability and resource efficiency over state-of-the-art ensemble attacks, including those rely on complex, optimized ensemble strategies.
    
\end{itemize}

\section{Related Work}
\label{sec:rel}

Various methods have been proposed to improve the transferability of adversarial examples, following similar principles in improving the generalizability of neural networks.

\noindent\textbf{Model ensemble attacks.} Model ensemble attacks generally combine outputs from multiple models. Specifically,~\citet{liu2017Delving} averages loss from multiple models.
~\citet{Dong2018momentum} further proposes logits or predicted probabilities ensemble. Advanced ensemble strategies are also studied.
Specifically,~\citet{Xiong2021SVRE} notices differences between ensemble models and proposes Stochastic Variance Reduction Ensemble (SVRE) attack to stabilize the gradient update.
~\citet{chen2023AdaEA} proposes an Adaptive Ensemble Attack (AdaEA), which adapts the models according to their specific contributions.
~\citet{tang2024ensemble} introduces a novel method of Stochastic Mini-batch black-box attack with Ensemble Reweighing (SMER) to boost the transferability.
~\citet{chen2024ensemble} uncovers two common weaknesses of model ensembles and relies on them to design the Common Weakness Attack (CWA). 
Although model ensemble attacks notably boost adversarial transferability, their practical utility is constrained by the substantial computational resources and time required for simultaneous gradient calculations across multiple models.

\noindent\textbf{Gradient optimization transfer techniques.} MI~\cite{Dong2018momentum} introduces momentum into adversarial attacks, stabilizing gradient update directions. NI~\cite{Lin2020Nesterov} further integrates Nesterov accelerated gradient into transfer-based attacks to better handle the history and future gradients. VT~\cite{Wang2021Variance} utilizes variance tuning to find a more stable direction for gradient updates. PGN~\cite{Ge2023flat} relied on penalizing gradient norm to achieve flat local maxima.

\noindent\textbf{Input transformation transfer techniques.} DI~\cite{Xie2019diversity} applies random resizing and padding operations to the input during each attack iteration to increase sample diversity and mitigate overfitting. TI~\cite{Dong2019Evading} optimizes adversarial perturbations using a set of translated images and also reduces computational costs by convolving gradients of untranslated images with a kernel matrix. SI~\cite{Lin2020Nesterov} follows a scale-invariant strategy, which optimizes the adversarial perturbations over the scale copies of the input image. SSA~\cite{long2022Frequency} transforms input images into the frequency domain and uses multiple input copies with diverse frequency features.

In this work, following the common practice in model ensemble attacks \cite{Xiong2021SVRE,chen2024ensemble}, we consider integrating the above gradient optimization and input transformation transfer techniques. Note that another type of transfer technique, model refinement~\cite{wu2020skip,zhao2025revisiting}, is excluded since it focuses on model modification rather than model ensemble.

\section{Methodology}
\label{sec:method}

\subsection{Preliminaries of Model Ensemble Attacks}

Given a classifier $f$ and benign input $x$ with ground-truth label $y$, $f(x)$ represents the logits output of classifier $f$ for input $x$. Naturally, the objective of adversarial attacks is to maximize $J(f({{x}^{adv}}),y)$ to craft an adversarial example ${x}^{adv}$ that is visually indistinguishable from $x$ but misleads the classifier to produce an incorrect output. Here, $J(f({{x}^{adv}}),y)$ is the loss function of the classifier, typically the cross-entropy loss function. Hence, adversarial example generation can be formalized as:
\begin{equation}
\underset{{{x}^{adv}}}{\mathop{\arg \max }}\,J(f({{x}^{adv}}),y), s.t.||{{x}^{adv}}-x|{{|}_{\infty }}\leq \varepsilon,
\label{eq1}
\end{equation}
\vskip -0.06in
\noindent where $\varepsilon$ denotes the constraint on adversarial perturbations using the $L_{\infty}$ norm. Instead of directly solving the complex optimization problem of Equation~\ref{eq1}, the Fast Gradient Sign Method (FGSM)~\cite{Goodfellow2015Explaining} and its iterative version, I-FGSM~\cite{Kurakin2018physical}, adopt the projected gradient descent.

The model ensemble is commonly used to improve adversarial transferability based on the assumption that adversarial directions learned from a diverse set of surrogate models can better generalize/transfer to unknown target models.
Early studies on model ensemble attacks~\cite{liu2017Delving} propose to average the loss values of $J(f({{x}^{adv}}),y)$ in Equation~\ref{eq2}, resulting in the ensemble loss function as:
\begin{equation}
\underset{{{x}^{adv}}}{\mathop{\arg \max }}\,\sum\nolimits_{i=1}^{k}{\frac{1}{k}J({{f}_{i}}({{x}^{adv}}),y)},
\label{eq2}
\end{equation}
\vskip -0.06in
\noindent where $k$ denotes the number of ensemble models.
Differently, logit output values can be averaged:
\begin{equation}
\underset{{{x}^{adv}}}{\mathop{\arg \max }}\,\sum\nolimits_{i=1}^{k}{J(\frac{1}{k}{{f}_{i}}({{x}^{adv}}),y)}.
\label{eq3}
\end{equation}
\vskip -0.06in

In this paper, we adopt the logit ensemble variant due to its simplicity and effectiveness \cite{Dong2018momentum}.

\subsection{Decoupling Within-iteration and Cross-iteration Model Diversity}
\label{sec:motiva}

\begin{figure}[t]
  \vskip -0.15in
  \centering
    \includegraphics[width=0.96\linewidth]{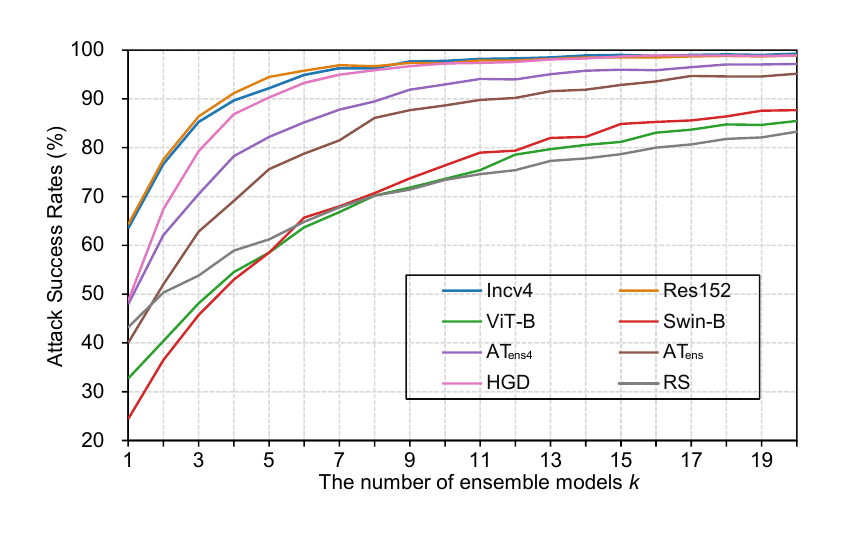}
     \vspace{-6pt}
    \caption{Attack success rates (\%) of MI transferring from an ensemble of $k$ models to eight different target models. The result for TI in the Appendix \ref{sec:addit-exp} shows similar patterns.}
    \label{fig:moti}
    \vskip -0.2in
\end{figure}

Model ensemble attacks are known to yield improved transferability but require significant time and memory consumption due to the gradient computations on multiple models. Consequently, existing work has commonly restricted the number of models (typically $<5$) for an acceptable resource efficiency~\cite{Ge2023flat,chen2023AdaEA}.
To fully investigate the impact of the number of models on the attack transferability, we conduct exploratory experiments by using 20 different models (see detailed descriptions in Section \ref{sec:set}).
As can be seen from the experimental results in Figure~\ref{fig:moti}, using more models consistently improves the transferability, despite the performance saturation when too many models are adopted.
This finding is supported by~\cite{hao2023TSEA,liu2024scaling,yao2025understanding}. Specifically, as demonstrated in \cite{hao2023TSEA}, increasing the number of models in adversarial attacks can reduce the upper bound of generalization error in empirical risk minimization. 

The above finding suggests that resource efficiency, which requires \textit{decreasing} the number of models, may contradict transferability, which requires \textit{increasing} the number of models.
However, we point out that this contradiction is caused by the common practice that the models used in each iteration should be identical.
To this end, we introduce the concept of cross-iteration and within-iteration model diversity.
These two dimensions of model diversity have distinct effects on transferability and resource efficiency.
On the one hand, both cross-iteration and within-iteration model diversity account for higher transferability.
On the other hand, resource efficiency is only related to the number of models per iteration, corresponding to within-iteration model diversity.
We present their formal definitions as follows.

\noindent\textbf{Definition 1} \noindent\textbf{\textit{Cross-iteration and Within-iteration Model Diversity}}. \textit{Given an iterative attack, the model diversity achieved by the number of distinct models with different architectures used throughout the entire process, $n$, is termed cross-iteration model diversity, denoted by $D_c$. The model diversity achieved by the number of distinct models used per iteration, $m$, is termed within-iteration model diversity, denoted by $D_i$. Here $m$ remains a constant in each iteration and $m\leq n$.}

\noindent\textbf{Quantifying diversity}. To explore the relation between the two dimensions of diversity in \textbf{Definition 1} and actual transferability, we further quantify the diversity.
Inspired by existing work on transferable attacks/defenses, we quantify them based on gradient similarity~\cite{kariyappa2019improving,li2020towards}.
Intuitively, different models yield different gradient directions on the input image.
In an ensemble attack, collective gradient directions from different models enable adversarial examples to better traverse the decision boundaries of multiple models, thereby improving the possibility of transferring them to unknown target models.
For within-iteration diversity, we compute the cosine similarity $S$ between gradients from different models in the same iteration:
\begin{equation}
D_i=\frac{2}{m(m-1)}\sum_{1\leq i < j\leq m}S(\nabla_x J(f_i(x)),\nabla_x J(f_j(x))).
\end{equation}
For cross-iteration diversity, we compute the cosine similarity between gradients from different models in two adjacent iterations:
\begin{equation}
D_c=
S(\sum\nolimits_{i=1}^{m}{\frac{\nabla_xJ({{f}_{i}}({x_t}))}{m}},\sum\nolimits_{i=1}^{m}{\frac{\nabla_xJ({{f}_{i}}({x_{t+1}}))}{m}}).
\end{equation}

\begin{table}[tbp]
\centering
\caption{Attack success rates (\%) of MI when varying the number of cross-iteration models ($n$) and/or within-iteration models ($m$).}
\label{tab:assum}
\resizebox{0.93\linewidth}{!}{
\begin{tabular}{cc|ccccc}
\toprule[1pt]
$n$ & $m$ & AT$_{ens4}$ & AT$_{ens}$ & HGD & RS & Avg.\\
\midrule[1pt]
3 & 3 & 37.6 & 23.5 & 27.0 & 34.9 & 30.8 \\
4 & 3 & \textbf{41.9} & 27.6 & 34.2 & 36.7 & 35.1 \\
4 & 4 & \textbf{41.9} & \textbf{27.7} & \textbf{34.3} & \textbf{36.9} & \textbf{35.2} \\
\bottomrule[1pt]
\end{tabular}}
\end{table}

\begin{table}[tbp]
\centering
\caption{Cross-iteration ($D_c$) and within-iteration ($D_i$) model diversity regarding gradient similarity for three attacks in Table~\ref{tab:assum}.}
\label{tab:sim}
\resizebox{0.93\linewidth}{!}{
\begin{tabular}{c|ccc|cc}
\toprule[1pt]
\multirow{2}{*}{$n$} & \multicolumn{2}{c}{$D_c$} & \multirow{2}{*}{$m$} & \multicolumn{2}{c}{$D_i$} \\
\cmidrule(lr){2-3} \cmidrule(lr){5-6}
& grad & sign(grad) & & grad & sign(grad) \\
\midrule[1pt]
3  & 0.0023 & 0.0021 & 3  & 0.0062 & 0.0061 \\
4  & \textbf{0.0022} & \textbf{0.0020}  & 4 & \textbf{0.0055} & \textbf{0.0053} \\
\bottomrule[1pt]
\end{tabular}}
\end{table}

Table~\ref{tab:assum} shows the transferability results when varying the number of within-iteration models ($m$) and/or cross-iteration models ($n$). 
We can observe that solely increasing $n$ from 3 to 4 largely boosts the transferability.
However, when further increasing $m$ from 3 to 4, the transferability is just improved a bit.
This difference indicates the greater power of using as many models as possible across iterations (without sacrificing the resource efficiency).
As expected, the diversity of the above three attacks quantified by gradient similarity in Table~\ref{tab:sim} is consistent with the results in Table~\ref{tab:assum}.

\subsection{Selective Ensemble Attack (SEA)}
\label{sec:SEA}

Conventional model ensemble attacks with identical models per iteration enforce $n=m$ in \textbf{Definition 1}.
As a result, even when there are many easily available models, they cannot use them to increase $D_c$ without increasing $D_i$ (i.e., requiring more resources).
To address this limitation, we propose the Selective Ensemble Attack (SEA).
SEA dynamically selects \textit{diverse} models across iterations to leverage as many models as we want to increase $D_c$ while maintaining $D_i$ (i.e., maintaining the resource efficiency).
SEA leads to overall higher diversity than the conventional ensemble methods, and as a result, improves the transferability.

Specifically, given $s$ easily accessible pre-trained models, denoted as $\mathcal{F}_s$ = \{$f_1$, $f_2$, $f_i$, ..., $f_s$\}, SEA($s$,$m$) dynamically selects $m$ models in each iteration according to the required resource constraint.
A larger $s$ generally leads to a larger number of cross-iteration models $n$, but $n$ may be smaller than $s$ when the iteration number is constrained (see detailed analyses of the expectation of $n$ in Appendix \ref{sec:expectation}).
In contrast, the conventional method Ens($s$,$m$) has to limit $n$ to $m$.
For model selection, we adopt the simple random selection and compare different sampling strategies in Table \ref{tab:samp_stra} of Section \ref{sec:ablation}.
The implementation details of SEA are presented in Algorithm \ref{alg:algorithm}.

\begin{algorithm}[t]
\caption{Selective Ensemble Attack (SEA)}
\label{alg:algorithm}
\textbf{Input}: A clean example $x$ with ground-truth label $y$; a classifier $f$ with loss function $J$; the logits of $m$ within-iteration models $l_1, l_2, ..., l_m$; The set of available models across iterations $\mathcal{F}_s=\{f_1, f_2,..., f_s\}$.\\
\textbf{Parameter}: Perturbation size $\varepsilon$; iteration number $T$ and decay factor $\mu$.\\
\textbf{Output}: Adversarial example $x^{adv}$.

\begin{algorithmic}[1] 
\STATE $\alpha=\varepsilon/T$; $x=x_0^{adv}$; $g_0=0$;
\FOR{$t=0$ to $T-1$}
\STATE Randomly select $m$ models from set $\mathcal{F}$:\\  ${\mathcal{F}}_{m}$ = $\{f_1, f_2,..., f_m\}$ = $random.sample(\mathcal{F}_s,m)$;
\STATE Input $x$ and output ${{l}_{i}}(x_{t}^{adv})$ for $i=1, 2, ..., m$;
\STATE Fuse the logits as $l(x_{t}^{adv})=\sum\nolimits_{i=1}^{m}{{{l}_{i}}(x_{t}^{adv})}/m$;
\STATE Get gradient as $g={{\nabla }_{x}}J(l(x_{t}^{adv}),y,\theta )$
\STATE Update ${{g}_{t+1}}$ by ${{g}_{t+1}}=\mu \cdot {{g}_{t}}+\frac{g}{||g|{{|}_{1}}}$;
\STATE Update $x_{t+1}^{adv}$ by $x_{t+1}^{adv}$ = $\text{Clip}_{x}^{\varepsilon }\{x_{t}^{adv}+\alpha \cdot \text{sign}({{g}_{t+1}})\}$;
\ENDFOR  
\STATE \textbf{return} ${{x}^{adv}}=x_{T}^{adv}$
\end{algorithmic}
\end{algorithm}

\section{Experiments}
\label{sec:exp}

\subsection{Experimental Setups} \label{sec:set}

\noindent\textbf{Dataset and models.} The experiments are conducted on the ImageNet-compatible dataset \cite{Russakovsky2015imagenet}, which is widely used for evaluating adversarial examples \cite{Dong2018momentum,Lin2020Nesterov}. This dataset consists of 1000 images from 1000 ImageNet categories (one image per category), which can be correctly classified by all the adopted models described as follows.

\begin{figure*}[htbp]
    \centering
    \includegraphics[width=0.98\textwidth]{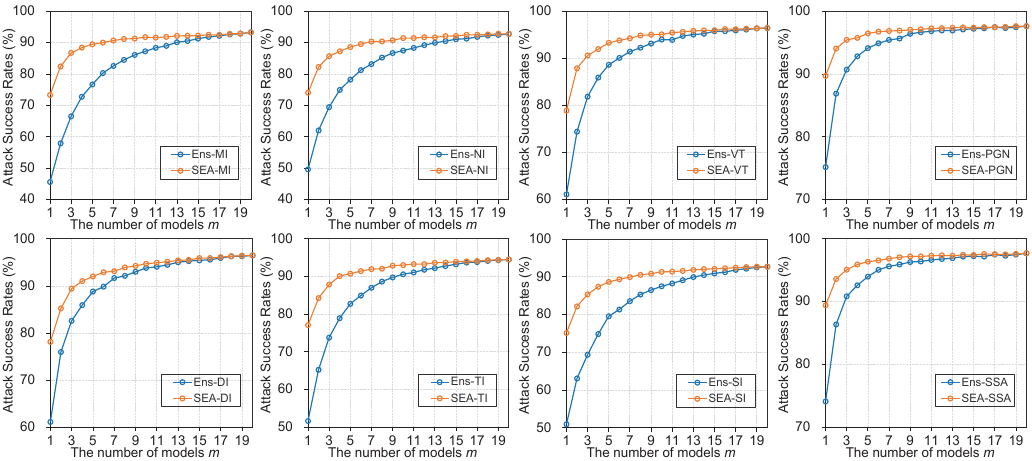}
    \vspace{-0.2cm}
    \caption{Attack success rate (\%) of Ens(20,$m$) vs. SEA(20,$m$) when within-iteration model $m$ varies from 1 to 20. Results are reported for eight transfer baselines and averaged over eight target models.}
    \label{fig:effi}
\end{figure*}
\begin{table*}[htbp]
\centering
\caption{Attack success rates (\%) of Ens vs. SEA with eight different transfer baselines. The upper bound using all 20 accessible models in each iteration is denoted as Ens(20,20), which costs $\times5$ more computational time and memory than Ens(20,4)/SEA(20,4).}
\label{tab:20-4}
\vspace{-0.5em}
\fontsize{6.5}{6.8}\selectfont
\setlength{\tabcolsep}{1.88mm}
\resizebox{0.95\linewidth}{!}{
\begin{tabular}{cccccccccccc}
\toprule[0.7pt]
\multirow{2}{*}{Baseline}& \multirow{2}{*}{Attack} & \multicolumn{2}{c}{CNNs} & \multicolumn{2}{c}{ViTs} & \multicolumn{4}{c}{Defenses} & \multirow{2}{*}{Avg.} \\ \cmidrule(lr){3-4} \cmidrule(lr){5-6} \cmidrule(lr){7-10} 
 &  & Incv4 & Res152 & ViT-B & Swin-B & AT$_{ens4}$ & AT$_{ens}$ & HGD & RS  \\ 
\toprule[0.7pt]
                           & \textcolor{gray}{Ens(20,20)}                                      & \textcolor{gray}{99.3} & \textcolor{gray}{98.9} & \textcolor{gray}{85.5} & \textcolor{gray}{87.4} & \textcolor{gray}{97.2} & \textcolor{gray}{95.2} & \textcolor{gray}{98.9} & \textcolor{gray}{82.9} & \textcolor{gray}{93.2} \\
                           & Ens(20,4)                      & 89.7                        & 91.2                        & 54.7                        & 52.0                        & 78.3                        & 69.1                        & 86.4                        & 59.6                        & 72.6                        \\
\multirow{-3}{*}{MI}       & SEA(20,4)                 & \textbf{97.7}               & \textbf{97.4}               & \textbf{77.3}               & \textbf{78.6}               & \textbf{93.6}               & \textbf{89.6}               & \textbf{96.5}               & \textbf{76.4}               & \textbf{88.4}               \\ \specialrule{0.04em}{1pt}{1pt}
                           & \textcolor{gray}{Ens(20,20)}                                      & \textcolor{gray}{99.4} & \textcolor{gray}{98.9} & \textcolor{gray}{84.7} & \textcolor{gray}{86.3} & \textcolor{gray}{96.5} & \textcolor{gray}{93.5} & \textcolor{gray}{99.1} & \textcolor{gray}{83.6} & \textcolor{gray}{92.8} \\
                           & Ens(20,4)                      & 92.1                        & 93.6                        & 57.4                        & 56.5                        & 78.3                        & 70.4                        & 88.6                        & 62.1                        & 74.9                        \\
\multirow{-3}{*}{NI}       & SEA(20,4)                 & \textbf{98.9}               & \textbf{97.8}               & \textbf{73.7}               & \textbf{76.1}               & \textbf{91.8}               & \textbf{87.9}               & \textbf{97.7}               & \textbf{73.1}               & \textbf{87.1}               \\ \specialrule{0.04em}{1pt}{1pt}
                           & \textcolor{gray}{Ens(20,20)}                                      & \textcolor{gray}{99.4} & \textcolor{gray}{99.2} & \textcolor{gray}{92.4} & \textcolor{gray}{93.0} & \textcolor{gray}{97.4} & \textcolor{gray}{96.5} & \textcolor{gray}{99.2} & \textcolor{gray}{93.8} & \textcolor{gray}{96.4} \\
                           & Ens(20,4)                      & 95.3                        & 95.1                        & 75.0                        & 73.6                        & 89.2                        & 85.4                        & 94.6                        & 79.3                        & 85.9                        \\ 
\multirow{-3}{*}{VT}       & SEA(20,4)                 & \textbf{97.5}               & \textbf{97.3}               & \textbf{84.5}               & \textbf{86.1}               & \textbf{94.6}               & \textbf{92.6}               & \textbf{97.0}               & \textbf{85.5}               & \textbf{91.9}               \\ \specialrule{0.04em}{1pt}{1pt}
                           & \textcolor{gray}{Ens(20,20)}                                      & \textcolor{gray}{98.9} & \textcolor{gray}{99.2} & \textcolor{gray}{95.6} & \textcolor{gray}{93.2} & \textcolor{gray}{98.5} & \textcolor{gray}{97.9} & \textcolor{gray}{99.3} & \textcolor{gray}{98.2} & \textcolor{gray}{97.6} \\
                           & Ens(20,4)                      & 97.5                        & 97.2                        & 87.4                        & 82.0                        & 95.2                        & 93.7                        & 96.3                        & 93.5                        & 92.9                        \\
\multirow{-3}{*}{PGN}      & SEA(20,4)                 & \textbf{98.2}               & \textbf{98.1}               & \textbf{92.3}               & \textbf{90.0}               & \textbf{97.0}               & \textbf{96.4}               & \textbf{98.3}               & \textbf{96.5}               & \textbf{95.9}               \\ \specialrule{0.04em}{1pt}{1pt}
                           & \textcolor{gray}{Ens(20,20)}                                      & \textcolor{gray}{99.4} & \textcolor{gray}{99.4} & \textcolor{gray}{92.9} & \textcolor{gray}{91.6} & \textcolor{gray}{98.1} & \textcolor{gray}{97.6} & \textcolor{gray}{99.3} & \textcolor{gray}{95.9} & \textcolor{gray}{96.8} \\
                           & Ens(20,4)                      & 96.5                        & 95.5                        & 74.9                        & 68.5                        & 91.7                        & 88.2                        & 95.9                        & 80.9                        & 86.5                        \\
\multirow{-3}{*}{DI}       & SEA(20,4)                 & \textbf{98.0}               & \textbf{97.3}               & \textbf{80.9}               & \textbf{77.2}               & \textbf{95.2}               & \textbf{93.0}               & \textbf{97.0}               & \textbf{85.8}               & \textbf{90.6}               \\ \specialrule{0.04em}{1pt}{1pt}
                           & \textcolor{gray}{Ens(20,20)}                                      & \textcolor{gray}{99.1} & \textcolor{gray}{98.6} & \textcolor{gray}{88.1} & \textcolor{gray}{82.1} & \textcolor{gray}{97.3} & \textcolor{gray}{96.6} & \textcolor{gray}{98.8} & \textcolor{gray}{95.3} & \textcolor{gray}{94.5} \\
                           & Ens(20,4)                      & 90.5                        & 91.3                        & 64.9                        & 51.0                        & 84.9                        & 79.6                        & 89.7                        & 80.0                        & 79.0                        \\
\multirow{-3}{*}{TI}       & SEA(20,4)                 & \textbf{97.0}               & \textbf{96.0}               & \textbf{81.1}               & \textbf{73.1}               & \textbf{94.1}               & \textbf{92.2}               & \textbf{96.3}               & \textbf{90.6}               & \textbf{90.1}               \\ \specialrule{0.04em}{1pt}{1pt}
                           & \textcolor{gray}{Ens(20,20)}                                      & \textcolor{gray}{99.6} & \textcolor{gray}{98.9} & \textcolor{gray}{84.1} & \textcolor{gray}{86.2} & \textcolor{gray}{96.1} & \textcolor{gray}{94.0} & \textcolor{gray}{99.2} & \textcolor{gray}{83.5} & \textcolor{gray}{92.7} \\
                           & Ens(20,4)                      & 91.2                        & 93.0                        & 58.1                        & 55.5                        & 77.6                        & 71.9                        & 88.2                        & 63.4                        & 74.9                        \\
\multirow{-3}{*}{SI}       & SEA(20,4)                 & \textbf{98.4}               & \textbf{98.2}               & \textbf{75.0}               & \textbf{75.8}               & \textbf{93.2}               & \textbf{87.8}               & \textbf{97.5}               & \textbf{72.9}               & \textbf{87.4}               \\ \specialrule{0.04em}{1pt}{1pt}
                           & \textcolor{gray}{Ens(20,20)}                                      & \textcolor{gray}{99.6} & \textcolor{gray}{99.5} & \textcolor{gray}{95.0} & \textcolor{gray}{93.7} & \textcolor{gray}{98.9} & \textcolor{gray}{98.0} & \textcolor{gray}{99.6} & \textcolor{gray}{97.2} & \textcolor{gray}{97.7} \\
                           & Ens(20,4)                      & 98.6                        & 98.2                        & 85.2                        & 81.2                        & 95.3                        & 94.5                        & 97.1                        & 90.6                        & 92.6                        \\
\multirow{-3}{*}{SSA}      & SEA(20,4)                 & \textbf{99.1}               & \textbf{98.8}               & \textbf{92.1}               & \textbf{90.4}               & \textbf{97.5}               & \textbf{96.4}               & \textbf{99.1}               & \textbf{93.8}               & \textbf{95.9}              \\
\bottomrule[0.7pt]
\end{tabular}
}
\end{table*}

For the target model, we consider 2 CNNs, 2 ViTs, and 4 Defenses. Specifically, 2 CNNs are Inception-v4 (Incv4) \cite{Szegedy2016inc-v3} and Resnet-v2-152 (Res152) \cite{He2016residual}. 2 ViTs are ViT-Base (ViT-B) \cite{dosovitskiy2020image} and Swin-Base (Swin-B) \cite{liu2021swin}. 4 Defenses are ens4-adv-Inception-v3 (AT$_{ens4}$), ens-adv-Inception-ResNet-v2 (AT$_{ens}$) \cite{Florian2018EnsembleAT}, HGD \cite{liao2018HGD}, and RS \cite{cohen2019RS}.
For the surrogate models, we select them from 20 pre-trained PyTorch models with 8 diverse architectures: resnet34/50/101/152 \cite{He2016residual}, vgg11/13/16/19 \cite{Simonyan2015vgg}, densenet121/161/169/201 \cite{huang2017densely}, fbresnet152 \cite{He2016residual}, dpn98/107/131 \cite{Chen2017dpn}, nasnetamobile \cite{zoph2018lnasnet}, senet154, and se\_resnet101/152 \cite{hu2018squeeze}. 

\noindent\textbf{Transfer baselines and hyper-parameter settings.} We integrate conventional model ensemble attacks (Ens) \cite{Dong2018momentum} or our SEA with eight popular transfer techniques that are based on gradient optimization (i.e., MI \cite{Dong2018momentum}, NI \cite{Lin2020Nesterov}, VT \cite{Wang2021Variance}, and PGN \cite{Ge2023flat}) or input transformation (i.e., DI \cite{Xie2019diversity}, TI \cite{Dong2019Evading}, SI \cite{Lin2020Nesterov}, and SSA \cite{long2022Frequency}). 
We also consider more complex ensemble strategies, such as SVRE \cite{Xiong2021SVRE}, AdaEA \cite{chen2023AdaEA}, SMER \cite{tang2024ensemble}, and CWA \cite{chen2024ensemble}.
All the above transfer baselines have been introduced in Section \ref{sec:rel}.

For Ens and our SEA, the logit outputs of all models are averaged with equal weights.
We set the perturbation budget $\varepsilon=16$, the number of iterations $T=10$.
All integrated transfer techniques adopt the settings in their original papers. Specifically, for MI and NI, we set the decay factor $\mu=1.0$. For VT, we set the $\beta=1.5$ and the number of copies $N=20$. For PGN, we set the number of copies $N=20$, the balancing factor $\delta=0.5$, and the upper bound of neighborhood size $\zeta=3.0 \times \varepsilon$. For DI, we set the probability $p=0.5$. For TI, we set the kernel length $k=7$. For SI, we set the number of copies $N=5$.
For SSA, we set the tuning factor $\rho=0.5$ and the number of spectrum transformations $N=20$. 
For SVRE, AdaEA, SMER, and CWA, default settings are adopted.

\noindent\textbf{Evaluation metrics.} We adopt attack success rate (\%) to evaluate transferability and time consumption and memory usage to evaluate resource efficiency. The attack success rate measures the model's misclassification rate on the 1000 adversarial examples from the dataset. Time consumption measures the time consumed to craft the 1000 adversarial examples from the dataset. Memory usage measures the maximum amount of GPU memory used during the attack. For attack success rate, higher is better, while for time consumption and memory usage, lower is better.

For clarity, we denote SEA and Ens under the ($s$,$m$) scenario as SEA($s$,$m$) and Ens($s$,$m$), respectively, where $s$ is the number of all easily accessible models and $m$ is the number of within-iteration models.
When disregarding the architectural differences of models, the resource efficiency is directly proportional to $m$.
For example, SEA(4,3) and Ens(4,3) are regarded as costing the same resource, and SEA(4,3) costs $1.5\times$ resource of SEA(4,2).
We also report the actual time and memory costs in Tables \ref{tab:compara}, \ref{tab:target}, \ref{tab:diver}, \ref{tab:samp_stra} and in Appendix \ref{sec:addit-exp}, \ref{sec:ens-strategy}, \ref{sec:batchsize}.
Additionally, the analysis of model loading time for SEA and Ens is reported in Appendix \ref{sec:addi-time}.

\begin{table*}[htb]
\centering
\caption{Transferability and resource efficiency of SEA vs. advanced ensemble strategies for (20,4).}
\label{tab:compara}
\vskip -0.06in
\fontsize{7}{7.2}\selectfont
\setlength{\tabcolsep}{1.2mm}
\resizebox{0.95\linewidth}{!}{
\begin{tabular}{cccccccccccccc}
\toprule[0.75pt]
\multirow{2}{*}{Baseline}& \multirow{2}{*}{Attack} & \multicolumn{2}{c}{CNNs} & \multicolumn{2}{c}{ViTs} & \multicolumn{4}{c}{Defenses} & \multirow{2}{*}{Avg.} & \multirow{2}{*}{\begin{tabular}[c]{@{}c@{}}Time \\(min)\end{tabular}} & \multirow{2}{*}{\begin{tabular}[c]{@{}c@{}}Memory \\(GB)\end{tabular}} \\ \cmidrule(lr){3-4} \cmidrule(lr){5-6} \cmidrule(lr){7-10}
 &   & Incv4 & Res152 & ViT-B & Swin-B & AT$_{ens4}$ & AT$_{ens}$ & HGD & RS  \\ 
\toprule[0.75pt]
\multirow{5}{*}{MI} 
                    & Ens          & 89.7          & 91.2          & 54.7          & 52.0          & 78.3          & 69.1          & 86.4          & 59.6          & 72.6          & 6.0 & \textbf{14.0}         \\
                    & SVRE         & 94.5          & 94.5          & 64.7          & 62.5          & 84.7          & 78.0          & 93.1          & 66.3          & 79.8          & 36.8         & 15.5          \\
                    & AdaEA        & 91.0          & 91.3          & 56.6          & 53.5          & 78.4          & 70.6          & 87.7          & 61.8          & 73.8          & 88.7         & 19.1          \\
                    & SMER         & 95.3          & 95.4          & 65.3          & 61.0          & 85.6          & 79.2          & 92.6          & 68.8          & 80.4          & 92.9         & 21.9          \\
                    & CWA          & 94.9          & 94.8          & 55.8          & 47.9          & 75.9          & 65.7          & 87.7          & 63.3          & 73.2          & 9.1          & 20.1          \\
                    & \textbf{SEA} & \textbf{97.7} & \textbf{97.4} & \textbf{77.3} & \textbf{78.6} & \textbf{93.6} & \textbf{89.6} & \textbf{96.5} & \textbf{76.4} & \textbf{88.4} & \textbf{5.9} & 14.1 \\  \specialrule{0.01em}{1.1pt}{1.1pt} 
\multirow{5}{*}{TI} 
                    & Ens          & 90.5          & 91.3          & 64.9          & 51.0          & 84.9          & 79.6          & 89.7          & 80.0          & 79.0   & \textbf{5.9}        & 14.4           \\
                    & SVRE         & 93.2          & 92.3          & 70.7          & 54.6          & 88.3          & 84.0          & 92.9          & 83.9          & 82.5          & 37.0         & 15.6          \\
                    & AdaEA        & 90.3          & 90.2          & 65.0          & 51.1          & 84.7          & 79.0          & 88.7          & 77.9          & 78.3          & 88.9         & 19.0          \\
                    & SMER         & 92.1          & 91.6          & 69.6          & 49.6          & 87.7          & 82.7          & 90.5          & 85.2          & 81.1          & 92.1         & 22.0          \\
                    & CWA          & 85.8          & 86.3          & 50.4          & 34.1          & 68.9          & 59.2          & 77.8          & 58.2          & 65.1          & 9.2          & 19.8          \\
                    & \textbf{SEA} & \textbf{97.0} & \textbf{96.0} & \textbf{81.1} & \textbf{73.1} & \textbf{94.1} & \textbf{92.2} & \textbf{96.3} & \textbf{90.6} & \textbf{90.1} & \textbf{5.9} & \textbf{14.3}      \\ 

\bottomrule[0.75pt]
\end{tabular}
}
\end{table*}
\begin{table}[tbp]
\centering
    \caption{Attack success rates (\%) of Ens vs. SEA on two commercial Vision APIs and large vision-language models with 100 randomly selected images.} 
\label{tab:api_LVLMs}
\vskip -0.05in
\fontsize{7}{7}\selectfont
\setlength{\tabcolsep}{1.6mm}
\resizebox{0.97\linewidth}{!}{
\begin{tabular}{ccccc}
\toprule[0.75pt]
 \multirow{2}{*}{Attack} & \multicolumn{2}{c}{Vision APIs} & \multicolumn{2}{c}{LVLMs} \\
\cmidrule(lr){2-3} \cmidrule(lr){4-5}
 & Google & Baidu & Qwen & ChatGLM \\
\midrule
Ens(20,4) & 44 & 52 & 40 & 39 \\
SEA(20,4) & \textbf{58} & \textbf{67} & \textbf{52} & \textbf{50} \\
\bottomrule[0.75pt]
\end{tabular}}
\end{table}

\begin{table*}[htb]
\centering
\caption{Comparison of Ens vs. SEA on target attacks with DI-TI-MI transfer baseline.}
\vspace{-0.5em}
\fontsize{7}{7}\selectfont
\setlength{\tabcolsep}{1.5mm}
\resizebox{0.95\linewidth}{!}{
\begin{tabular}{clcccccccccc}
\toprule[0.7pt]
 \multirow{2}{*}{Attack} & \multicolumn{2}{c}{CNNs} & \multicolumn{2}{c}{ViTs} & \multicolumn{4}{c}{Defenses} & \multirow{2}{*}{Avg.} & \multirow{2}{*}{\begin{tabular}[c]{@{}c@{}}Time \\(min)\end{tabular}} & \multirow{2}{*}{\begin{tabular}[c]{@{}c@{}}Memory \\(GB)\end{tabular}} \\ \cmidrule(lr){2-3} \cmidrule(lr){4-5} \cmidrule(lr){6-9}
 & Incv4 & Res152 & ViT-B & Swin-B & AT$_{ens4}$ & AT$_{ens}$ & HGD & RS\\ 
\specialrule{0.06em}{1.2pt}{1.2pt} 
Ens(20,4) & 55.4 & 55.0 & 10.4 & 1.9 & 36.1 & 35.7 & 52.4 & 0.7 & 31.0 & 41.6 & 17.5 \\
SEA(20,4) & \textbf{96.3} & \textbf{94.5} & \textbf{37.6} & \textbf{15.0} & \textbf{85.4} & \textbf{85.6} & \textbf{89.9} & \textbf{6.7} & \textbf{63.9} & 40.7 & 17.9 \\
SEA(20,3) & 95.7 & 93.9 & 36.0 & 14.7 & 84.6 & 84.6 & 88.7 & 6.0 & 63.0 & \textbf{30.6} & \textbf{15.6} \\
\bottomrule[0.7pt]
\end{tabular}}
\label{tab:target}
\end{table*}

\begin{figure}[tbp]
     \centering
    \includegraphics[width=0.75\linewidth]{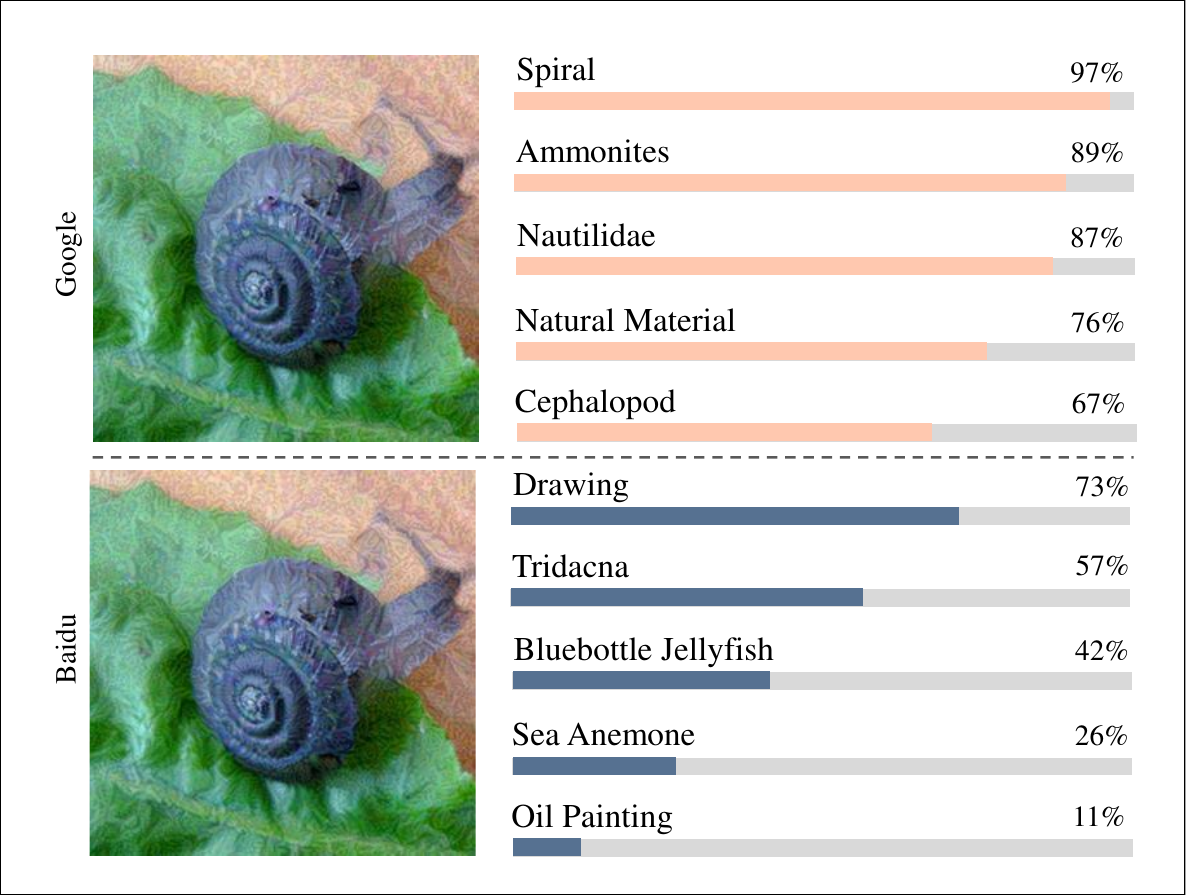}
    \captionof{figure}{SEA adversarial images on Google and Baidu Cloud Vision APIs. The ground truth label of the original image is ``snail''.}
    \label{fig:api}
    \vspace{-1.0em}
\end{figure}

\begin{table*}[tbp]
\begin{center}
\caption{Transferability and resource efficiency of Ens vs. SEA with similar or diverse surrogates. In the similar cases, model ensemble is resnet18/34/50/101, vgg11/13/16/19, densenet121/161/169/201, or dpn68/98/107/131. In the diverse case, model ensemble is resnet101/vgg11/densenet161/dpn131. MI is the transfer baseline. Results for TI in the Appendix \ref{sec:addit-exp} show similar patterns.}
\vspace{-0.3em}
\label{tab:diver}
\resizebox{0.95\linewidth}{!}{
\begin{tabular}{cccccccccccccc}
\toprule[1pt]
\multirow{2}{*}{Baseline}& \multirow{2}{*}{Attack} & \multicolumn{2}{c}{CNNs} & \multicolumn{2}{c}{ViTs} & \multicolumn{4}{c}{Defenses} & \multirow{2}{*}{Avg.} & \multirow{2}{*}{\begin{tabular}[c]{@{}c@{}}Time \\(min)\end{tabular}} & \multirow{2}{*}{\begin{tabular}[c]{@{}c@{}}Memory \\(GB)\end{tabular}} \\ \cmidrule(lr){3-4} \cmidrule(lr){5-6} \cmidrule(lr){7-10}
 &   & Incv4 & Res152 & ViT-B & Swin-B & AT$_{ens4}$ & AT$_{ens}$ & HGD & RS  \\ 
\toprule[1pt] 
                           & \textcolor{gray}{Ens(4,4)} & \textcolor{gray}{85.7} & \textcolor{gray}{91.6} & \textcolor{gray}{51.2} & \textcolor{gray}{51.3} & \textcolor{gray}{74.5} & \textcolor{gray}{63.5} & \textcolor{gray}{90.9} & \textcolor{gray}{62.0} & \textcolor{gray}{71.3} & \textcolor{gray}{1.8} & \textcolor{gray}{7.3}  \\
                           & Ens(4,3)        & 80.7                        & 86.9                        & 46.4                        & 42.6                        & 68.6                        & 55.3                        & 84.7                        & 58.1                        & 65.4                        & 1.5                        & 6.9                         \\
                           \multirow{-4}{*}{resnet} & SEA(4,3)    & \textbf{86.0}               & \textbf{91.7}               & \textbf{51.3}               & \textbf{49.2}               & \textbf{74.6}               & \textbf{63.0}               & \textbf{90.5}               & \textbf{61.3}               & \textbf{71.0}               & 1.5                        & 6.9                         \\
  \specialrule{0.06em}{1.2pt}{1.2pt}
                           & \textcolor{gray}{Ens(4,4)} & \textcolor{gray}{90.1} & \textcolor{gray}{81.0} & \textcolor{gray}{45.4} & \textcolor{gray}{43.8} & \textcolor{gray}{64.2} & \textcolor{gray}{50.9} & \textcolor{gray}{74.5} & \textcolor{gray}{52.3} & \textcolor{gray}{62.8} & \textcolor{gray}{2.6} & \textcolor{gray}{10.6} \\
                           & Ens(4,3)        & 86.3              & 76.6                        & 40.9                        & 38.6                        & 57.7                        & 43.9                        & 67.2                        & 50.4                        & 57.7                        & 2.0                        & 10.1                        \\
        \multirow{-3}{*}{vgg}                      & SEA(4,3)    & \textbf{89.1}               & \textbf{80.9}               & \textbf{44.6}               & \textbf{42.2}               & \textbf{63.3}               & \textbf{48.9}               & \textbf{74.8}               & \textbf{52.1}               & \textbf{62.0}               & 2.1                        & 9.8                         \\
   \specialrule{0.06em}{1.2pt}{1.2pt}
                           & \textcolor{gray}{Ens(4,4)} & \textcolor{gray}{91.3} & \textcolor{gray}{93.2} & \textcolor{gray}{58.3} & \textcolor{gray}{56.3} & \textcolor{gray}{78.6} & \textcolor{gray}{71.0} & \textcolor{gray}{91.7} & \textcolor{gray}{63.2} & \textcolor{gray}{75.5} & \textcolor{gray}{3.1} & \textcolor{gray}{16.8} \\
                           & Ens(4,3)        & 87.6                        & 90.0                        & 52.1                        & 51.3                        & 73.3                        & 66.1                        & 87.8                        & 60.6                        & 71.1                        & 2.4                        & 13.6                        \\
           \multirow{-3}{*}{densenet}                 & SEA(4,3)    & \textbf{91.1}               & \textbf{93.3}              & \textbf{57.5}               & \textbf{57.4}               & \textbf{78.9}               & \textbf{71.8}               & \textbf{90.3}               & \textbf{63.7}               & \textbf{75.5}               & 2.4                        & 13.8                        \\
\specialrule{0.06em}{1.2pt}{1.2pt}
                           & \textcolor{gray}{Ens(4,4)} & \textcolor{gray}{86.9} & \textcolor{gray}{90.2} & \textcolor{gray}{55.0} & \textcolor{gray}{54.8} & \textcolor{gray}{76.8} & \textcolor{gray}{73.3} & \textcolor{gray}{87.0} & \textcolor{gray}{57.5} & \textcolor{gray}{72.7} & \textcolor{gray}{6.4} & \textcolor{gray}{21.9} \\
                           & Ens(4,3)        & 84.6                        & 87.2                        & 50.5                        & 50.5                        & 73.1                        & 67.4                        & 83.5                        & 55.2                        & 69.0                        & 5.2                        & 20.3                        \\
                        \multirow{-3}{*}{dpn}        & SEA(4,3)    & \textbf{87.5}                       & \textbf{90.0}               & \textbf{53.1}               & \textbf{56.2}               & \textbf{78.6}               & \textbf{72.9}               & \textbf{87.9}               & \textbf{58.2}               & \textbf{73.1}               & 5.0                        & 20.2                        \\
\specialrule{0.06em}{1.2pt}{1.2pt}
                           & \textcolor{gray}{Ens(4,4)} & \textcolor{gray}{91.7} & \textcolor{gray}{93.6} & \textcolor{gray}{59.2} & \textcolor{gray}{60.4} & \textcolor{gray}{81.0} & \textcolor{gray}{75.8} & \textcolor{gray}{91.7} & \textcolor{gray}{63.2} & \textcolor{gray}{77.1} & \textcolor{gray}{4.5} & \textcolor{gray}{17.9} \\
                           & Ens(4,3)        & 88.5                        & 90.2                        & 51.9                        & 53.1                        & 75.6                        & 67.8                        & 86.1                        & 58.5                        & 71.5                        & 3.5                        & 16.0                        \\
                           \multirow{-3}{*}{diverse} & SEA(4,3)    & \textbf{92.0}               & \textbf{92.9}               & \textbf{58.0}               & \textbf{61.0}               & \textbf{80.1}               & \textbf{74.7}               & \textbf{91.3}               & \textbf{63.7}               & \textbf{76.7}               & 3.5                        & 16.1                        \\
\bottomrule[1pt] 
\end{tabular}
}

\end{center}
\end{table*}

\subsection{SEA in Various Transfer Scenarios}
\label{sec:Ens(20)}

Our SEA gives the potential to use many models without increasing the resource costs.
To explore the potential of utilizing many models in SEA, we assume easy access to 20 pre-trained models from the PyTorch library.
As can be seen from Figure~\ref{fig:effi}, when varying the within-iteration model $m$ from 1 to 20, our SEA(20,$m$) consistently surpasses Ens(20,$m$) in all cases.
The performance gap becomes gradually larger as $m$ decreases.
This trend indicates that the superiority of our SEA stands out especially when the resource is restricted in practice.
In particular, when the common $m=4$ is used, our SEA(20,4) outperforms Ens(20,4) by 8.5\% on average.
More detailed results for (20,4) on various target models are reported in Table~\ref{tab:20-4}, and results for (4,3), a more commonly used ensemble setting are in Table~\ref{tab:classi} of Appendix~\ref{sec:addit-exp}. 

We have the following observations.
First, SEA performs better on all diverse (black-box) target models including CNNs, ViTs, and Defenses.
This is because SEA involves more models across iterations (but without consuming more resources).
Second, Ens(4,3) and SEA(4,3) consume very close resources, because they have the same within-iteration models.
Third, Table~\ref{tab:classi} shows that different transfer baselines consume different costs, and more costs generally lead to more effective attacks.

\textbf{SEA vs. advanced ensemble strategies.} We further compare our SEA to more complex, advanced model ensemble strategies, such as SVRE, AdaEA, SMER, and CWA.
These strategies optimize the ensemble attack from different perspectives (see descriptions in Section~\ref{sec:rel}), beyond equally averaging the logit or loss values of surrogate models, and follow Ens to use identical models across iterations.
As can be seen from Table~\ref{tab:compara}, our SEA still achieves the best transferability.
For resource efficiency, the simplicity of the random model selection makes SEA consume substantially less computational time and memory.   
We also find that combining the above complex ensemble strategies with our SEA can further improve transferability. See detailed results in Table~\ref{tab:ensem} of Appendix~\ref{sec:ens-strategy}.

\noindent\textbf{Attacking real-world systems.} We further compare Ens and SEA in attacking real-world systems: the commercial vision APIs, Google and Baidu Cloud Vision, and the large vision-language models (LVLM), Qwen and ChatGLM. Notably, given an (original or adversarial) image, we prompt the LVLMs using "Output a single-word label in English for this image." An adversarial image is considered successful when its output label differs from that of the original image. 
Table \ref{tab:api_LVLMs} shows that SEA achieves better results than Ens on both APIs and LVLMs, with the MI transfer baseline. 
Figure~\ref{fig:api} further shows successful SEA image examples on both APIs. Additional visualizations, also on two LVLMs, are provided in the Appendix \ref{sec:visualization}, where for LVLMs, we also try a finer-grained testing with the prompt "Describe the image no more than 50 words".

\noindent\textbf{Targeted attacks.} We further compare SEA and Ens under more challenging targeted attacks. Specifically, we follow the common practice \cite{wei2023enhancing,zhao2021success}, using 100 iterations and a step size of 2, with DI-TI-MI as the baseline. The results in Table \ref{tab:target} demonstrate that SEA consistently outperforms Ens. For instance, SEA(20,3) achieves an 32.0\% improvement in average attack success rate while reducing resource consumption compared to Ens(20,4), highlighting the superiority of our method.

\subsection{Ablation Studies} \label{sec:ablation}

\begin{table}[tbp]
\centering
\vspace{-1.5em}
\caption{Attack success rates (\%), Time consumption (min), and Memory usage (GB) of SEA under different sampling strategies. SEA(20,2) is used.} 
\label{tab:samp_stra}
\fontsize{8}{7.5}\selectfont
\setlength{\tabcolsep}{1.0mm}
\resizebox{\linewidth}{!}{
\begin{tabular}{ccccc}
\toprule[0.8pt]
\multirow{2}{*}{Base}  & \multirow{2}{*}{\begin{tabular}[c]{@{}c@{}}Sampling\\  strategy\end{tabular}} & \multirow{2}{*}{\begin{tabular}[c]{@{}c@{}}ASR\\     (\%)\end{tabular}} & \multirow{2}{*}{\begin{tabular}[c]{@{}c@{}}Time\\      (min)\end{tabular}} & \multirow{2}{*}{\begin{tabular}[c]{@{}c@{}}Memory\\      (GB)\end{tabular}} \\ &     &     &    &      \\  \midrule
\multirow{4}{*}{MI}   & random w/ replacement          & 83.5     & \textbf{4.2}      & 10.3    \\ 
& random w/o replacement          & \textbf{85.5}   &  \textbf{4.2}    &   \textbf{10.2}  \\
& optimization    & 83.0              & 4.3               & 10.5             \\
& model scheduling & 84.1              & 48.6               & 12.7             \\ \midrule              
\multirow{4}{*}{TI}   & random w/ replacement           & 86.4     & 4.3      & 10.4    \\
& random w/o replacement         & \textbf{87.6}    & \textbf{4.2}     & \textbf{10.3}    \\
& optimization    & 85.9     & 4.4               & 10.7              \\
& model scheduling & 86.7              & 48.7               & 12.8    \\
\bottomrule[0.8pt]
\end{tabular}}
\end{table}

\noindent\textbf{Diversity of surrogate models.} 
To validate the general effectiveness of our SEA, we consider ensemble settings with similar or diverse surrogate models.
Specifically, we adopt four ensemble models from the \textit{resnet}, \textit{vgg}, \textit{densenet}, or \textit{dpn} family, as well as their combination, \textit{diverse}.
As can be seen from Table \ref{tab:diver}, our SEA(4,3) consistently achieves higher transferability than Ens(4,3) across different model combinations. 
In addition, using a more diverse combination of surrogate models yields generally higher transferability for both Ens and SEA.
We also validate challenging cases with the ensemble of CNNs and ViTs in Appendix \ref{sec:integrating}, showing that even with such highly diverse model architectures, there are no opposite effects~\cite{chen2023AdaEA}.

\begin{figure}[tbp]
\centering
    \includegraphics[width=0.95\linewidth]{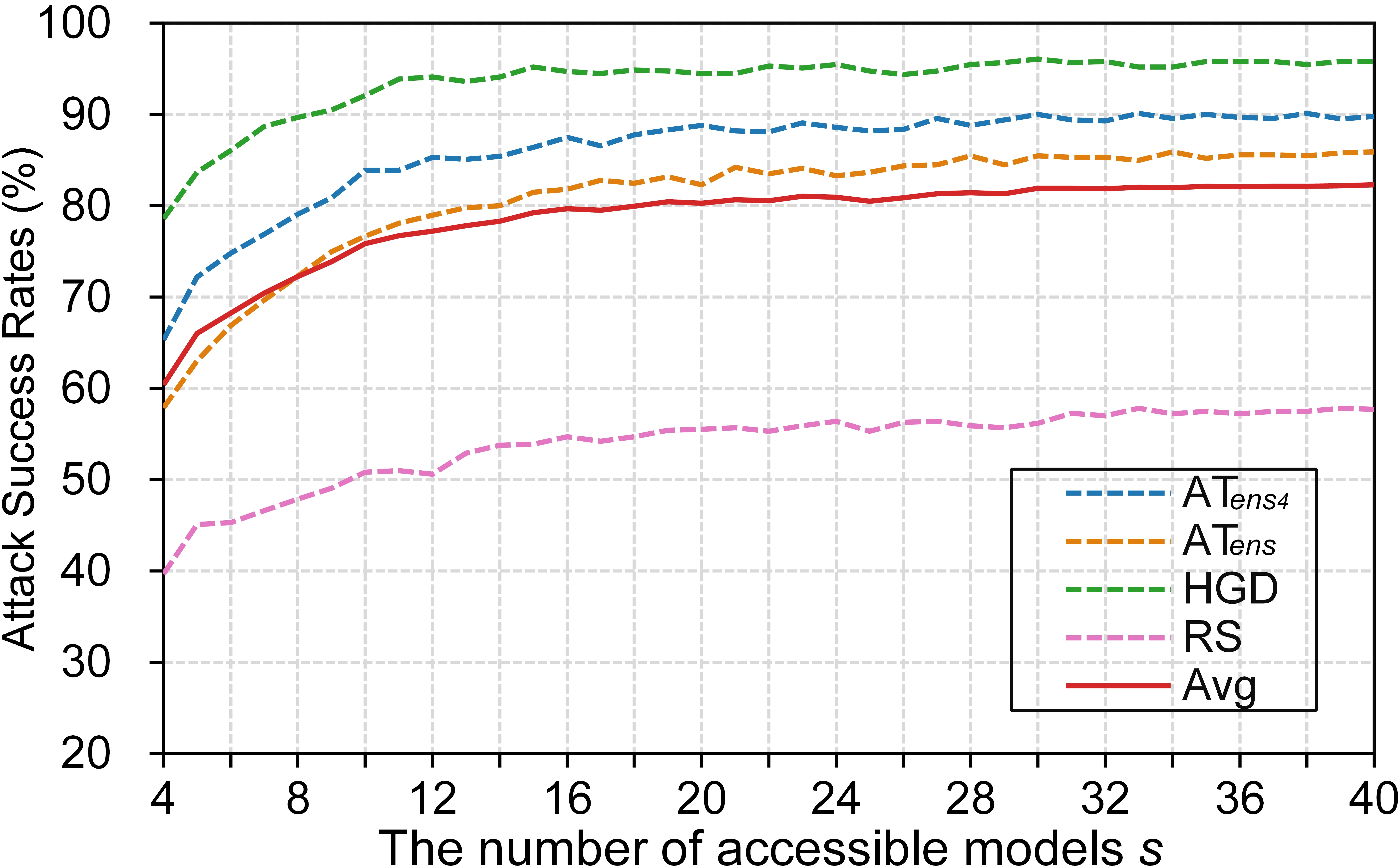}
    \vskip -0.07in 
    \captionof{figure}{Attack success rate (\%) of SEA ($s$, 4) by varying $s$ from 4 to 40.}
    \label{fig:vary-n}
\end{figure}

\noindent\textbf{Model selection strategy.}
Selecting a subset of models from all models in each iteration results in an enormous search space.
For example, when selecting 2 models from 20, there are $190^{10}$ possibilities over 10 iterations.
Here, we compare whether different selection strategies may have an impact on SEA.
Specifically, we compare three model selection strategies: 1) our random selection (with or without replacement across iterations); 2) an optimization-based selection \cite{zhu2024learning,liu2021direct}, which updates the probability of each model being selected through differential augmentation; 3) a model scheduling approach \cite{hu2022model}, which prioritizes models with representative decision boundaries (measured by gradient direction outlier degree) while progressively deprioritizing previously selected or defeated models across attack stages.
As can be seen from Table \ref{tab:samp_stra}, random sampling without replacement yields the best results.
This is because it avoids selecting the same models as those in previous iterations, leading to higher cross-iteration model diversity.
For example, for SEA(20,2), random selection without replacement for 10 iterations yields the number of cross-iteration models $n=s=20$.
The stability of random sampling is analyzed in Table \ref{tab:stable} of Appendix \ref{sec:Stability}.

\noindent\textbf{The number of accessible models $s$.}
Similar to Figure \ref{fig:effi}, we further conduct experiments on SEA(s,4) with varying $s$. Specifically, we fix $m=4$ and vary $s$ from 4 to 40 using the MI baseline to craft adversarial examples. As shown in Figure \ref{fig:vary-n}, the attack success rates steadily increase as $s$ increases. The growth is particularly notable when $s$ increases from 4 to 20.
After $s$ exceeds 20, the improvement in attack success rates becomes slower, which is attributed to a certain level of saturation in the ensemble attack performance, as demonstrated in Figures \ref{fig:moti} and \ref{fig:moti-TI}.

\vspace{-0.9em}
\section{Conclusion and Outlook}

In this work, we point out the limitation of the existing model ensemble attacks in achieving a good trade-off between transferability and resource efficiency regarding time consumption and memory usage. To address this limitation, we propose to drop the common practice that models used per iteration should be identical.
Concretely, we propose Selecting Ensemble Attack (SEA), which decouples the cross-iteration and within-iteration model diversity by dynamically selecting diverse models across iterations.
In this way, SEA can use as many models as we want to improve transferability without sacrificing resource efficiency.
Extensive experiments in various settings demonstrate the general superiority of our SEA. 
Overall, our SEA makes it possible to flexibly balance transferability and resource efficiency based on the actual resource constraint.

For future work, it is promising to further relax the constraint that the within-iteration models should be the same across iterations, possibly by dynamically selecting a different number of models per iteration.
This will introduce more flexibility into our SEA, leading to a finer-grained adjustment of transferability vs. resource efficiency.
It is also interesting to explore the generalizability of SEA to other domains beyond image classification.

{
    \small
    \bibliographystyle{ieeenat_fullname}
    \bibliography{main}
}

\clearpage
\appendix

\section{Analysis of the Expectation of $n$}
\label{sec:expectation}

\begin{figure}[htbp]
\begin{center}
\begin{minipage}{0.48\textwidth}
     \centering
    \includegraphics[width=\textwidth]{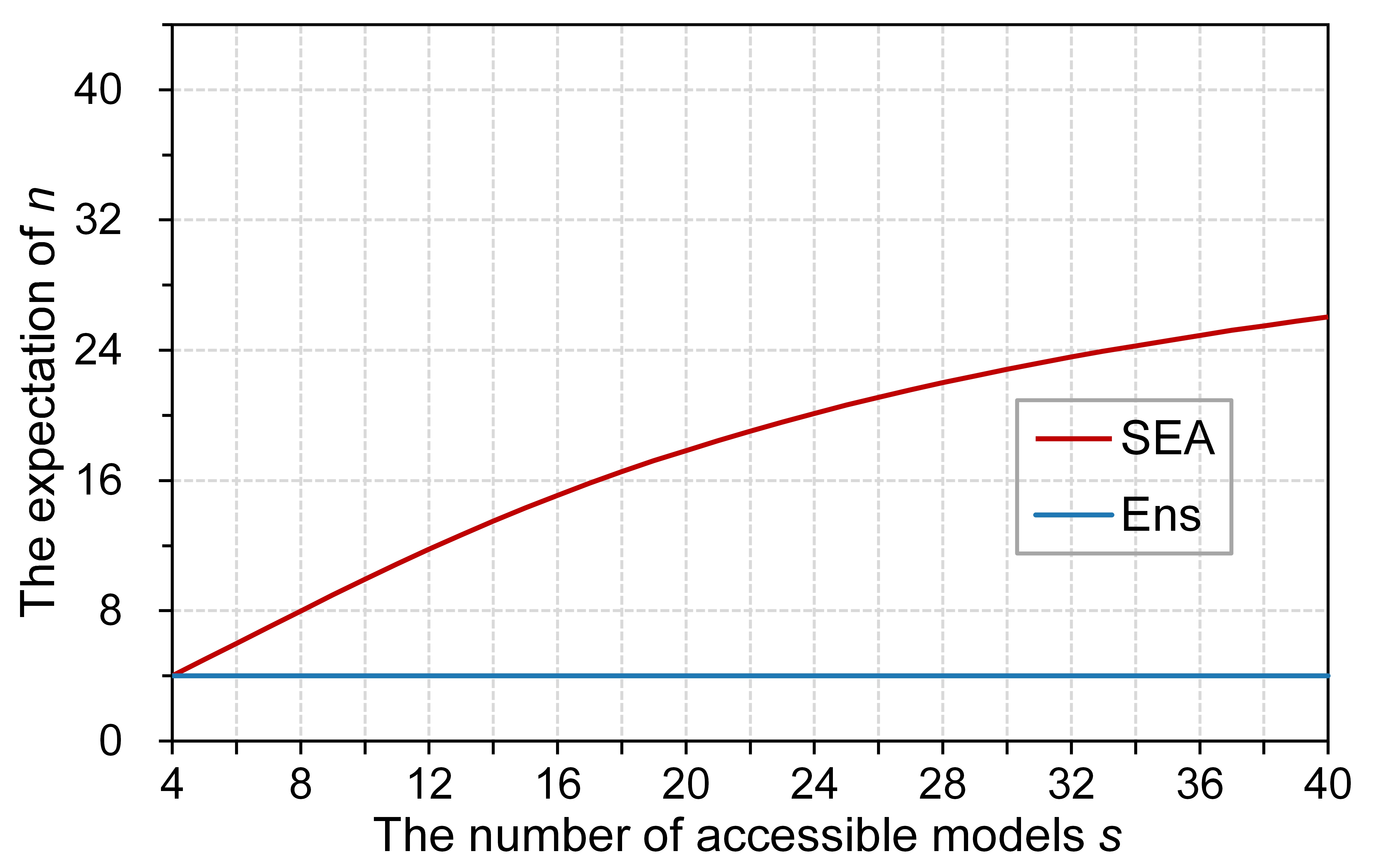}
    \caption{The expectation of $n$ of Ens vs. SEA by varying $s$ from 4 to 40. The number of within-iteration models $m$ is 4.}
    \label{fig:exp-vary-s}
\end{minipage} \hfill
\begin{minipage}{0.48\textwidth}
     \centering
    \includegraphics[width=\textwidth]{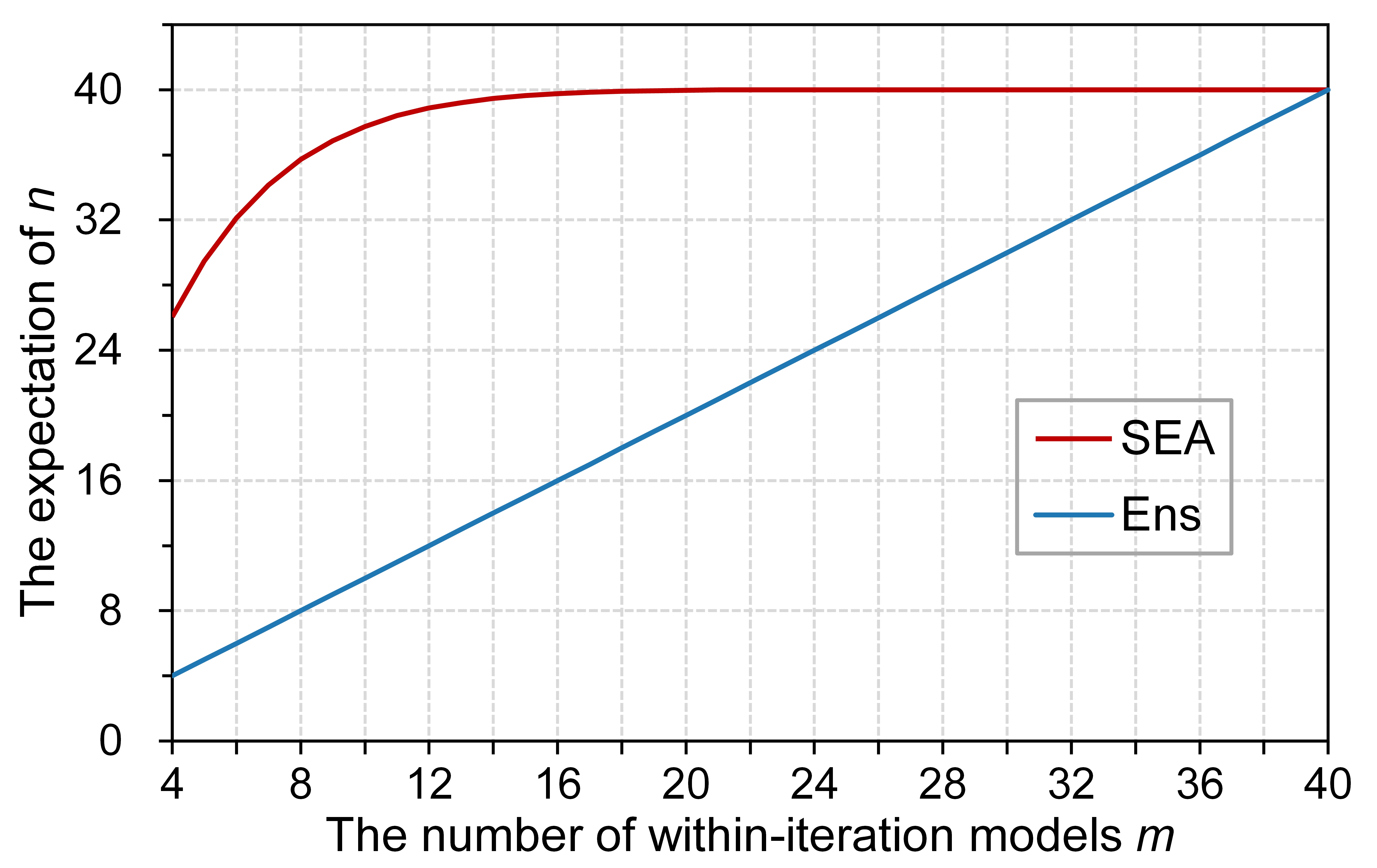}
    \caption{The expectation of $n$ of Ens vs. SEA by varying $m$ from 4 to 40. The number of accessible models $s$ is 40.}
    \label{fig:exp-vary-m}
\end{minipage}
\end{center}
\end{figure}

We further quantitatively analyze the expectation of distinct models across iterations ($n$) used by Ens and SEA under the same conditions. Specifically, given the number of all easily accessible models $s$, the number of within-iteration models $m$, and the total iterations $T$, the expectation of distinct models across iterations is given by: $s \cdot \left( 1 - \left( \frac{s - m}{s} \right)^T \right)$.
The derivation is as follows: For any given model, the probability of not being selected in a single iteration is $\frac{s - m}{s}$. Therefore, the probability that it is never selected over $T$ iterations is $\left( \frac{s - m}{s} \right)^T$, and the probability that it is selected at least once is $\left( 1 - \left( \frac{s - m}{s} \right)^T \right)$. By the linearity of expectation, the total expected number of distinct models selected over all iterations is the sum of the selection probabilities for each model, which yields: $s \cdot \left( 1 - \left( \frac{s - m}{s} \right)^T \right)$.
As shown in Figure \ref{fig:exp-vary-s} and Figure \ref{fig:exp-vary-m}, the expectation of $n$ in SEA increases as $s$ or $m$ increases.
In contrast, the expectation of $n$ for Ens is always limited to $m$ since the same models are reused in each iteration.

\section{Stability Testing of Random Sampling in SEA}
\label{sec:Stability}

\begin{table}[htbp]
\begin{center}
\caption{The results (avg±std) of repeating Ens and SEA 10 times with the MI baseline.} 
\label{tab:stable}
\vskip -0.1in
\resizebox{\linewidth}{!}{
\begin{tabular}{ccccc}
\toprule[0.9pt]
Attack & V19 &  D121 & ViT-B & Swin-B \\ \midrule[0.9pt]
Ens(4,3) & 73.9(±0.66) & 71.1(±0.82) & 26.3(±0.53) & 26.8(±0.98) \\ 
SEA(4,3) & 78.9(±0.61) & 77.4(±0.33) & 30.8(±0.41) & 32.6(±0.99) \\ 
SEA(4,2) & 78.4(±1.01) & 77.2(±0.78) & 30.1(±0.70) & 31.9(±0.81) \\
\bottomrule[0.9pt]
\end{tabular}}
\end{center}
\end{table}

We conduct an in-depth analysis of the stability of random sampling. Specifically, we repeat the experiments of SEA and Ens 10 times using the MI baseline. The experimental results in Table \ref{tab:stable} show that Ens(4,3) and SEA(4,3) exhibit comparable and low variability, indicating that the random sampling strategy offers good stability. It should be noted that the four-model ensemble employs Incv3 \cite{Szegedy2016inc-v3}, Incv4, IncResv2 \cite{Szegedy2017incRes}, and Res152 \cite{He2016residual} as surrogate models, consistent with prior works \cite{Lin2020Nesterov,Wang2021Variance}. To ensure a black-box evaluation, Vgg19 (V19) and Densenet121 (D121) are used as CNN test (target) models, while other test models remain unchanged.

\section{Model Loading Time} \label{sec:addi-time}

In this work, we load all available models at the beginning before the iterative attack optimization.
This is fair for Ens and SEA because, given $s$ accessible models, Ens would also try to optimize the set of used $m$ models. Specifically, Ens would run the experiments multiple times, each time with a different set of $m$ models, and select the best run based on the transferability (between surrogate models).
Table \ref{tab:stable} confirms that such pre-selection is valuable since the results do vary across different sets of models.
In contrast, our SEA does not require such a pre-selection but dynamically varies the set of models used in each iteration.
Compared to GPU memories consumed for gradient calculation, the 4 models in SEA(4,3) only occupy 0.96 GB of memory, representing 4.04\% and 1.22\% of the total memory capacity on 3090 (24 GB) and A800 (80 GB) GPUs, respectively. Similarly, 20 models occupy 20.2\% and 6.1\% of the total memory capacity on 3090 and A800, respectively.

Furthermore, the results in Table \ref{tab:addi-time} confirm that even with the additional time incurred by dynamic model loading, SEA still outperforms Ens in terms of attack effectiveness and efficiency in almost all cases when fewer models are used per iteration.
Specifically, we measure the model loading time of 20 models used in SEA (20,4), and the average time is 3.7s, with variations from 2.5s to 6.1s. For an ensemble of 4 models and 10 iterations, the additional time of SEA is $3.7\times4\times(10-1)= 133.2$ (s).

\begin{table}[tbp]
\begin{center}
\caption{The comparison between Ens(20,4) with SEA(20,2).} 
\label{tab:addi-time}
\fontsize{8}{8}\selectfont
\setlength{\tabcolsep}{1.1mm}
\resizebox{\linewidth}{!}{
\begin{tabular}{ccccc}
\toprule[0.75pt]
  Baseline        & Attack          & ASR(\%)           & Time(min)         & Memory(GB) \\ \midrule[0.75pt]
                  & Ens(20,4)       & 72.7              & 5.4               & 14.6               \\ 
 \multirow{-2}{*}{MI}           & SEA(20,2)       & \textbf{82.4}     & \textbf{5.3}      & \textbf{11.2}      \\\midrule
                  & Ens(20,4)       & 74.9              & 5.4               & 14.7               \\ 
  \multirow{-2}{*}{NI}            & SEA(20,2)       & \textbf{82.2}     & \textbf{5.3}      & \textbf{11.3}      \\ \midrule
                  & Ens(20,4)       & 85.9              & 113.7             & 14.3               \\ 
 \multirow{-2}{*}{VT}           & SEA(20,2)       & \textbf{87.9}     & \textbf{59.2}     & \textbf{11.6}      \\ \midrule
                  & Ens(20,4)       & 92.8              & 193.9             & 23.3               \\ 
  \multirow{-2}{*}{PGN}          & SEA(20,2)       & \textbf{94.0}     & \textbf{100.7}    & \textbf{19.7}      \\ \midrule
                  & Ens(20,4)       & \textbf{85.8}     & 5.5               & 14.3               \\ 
  \multirow{-2}{*}{DI}            & SEA(20,2)       & 85.2              & \textbf{5.3}      & \textbf{11.4}      \\ \midrule
                  & Ens(20,4)       & 78.9              & 5.5               & 14.3               \\ 
  \multirow{-2}{*}{TI}            & SEA(20,2)       & \textbf{84.3}     & \textbf{5.4}      & \textbf{11.5}      \\ \midrule
                  & Ens(20,4)       & 74.8              & 27.3              & 14.4               \\ 
  \multirow{-2}{*}{SI}            & SEA(20,2)       & \textbf{85.4}     & \textbf{16.1}     & \textbf{11.4}      \\ \midrule
                  & Ens(20,4)       & 92.6              & 113.0             & 15.8               \\ 
 \multirow{-2}{*}{SSA}           & SEA(20,2)       & \textbf{93.7}     & \textbf{60.1}     & \textbf{11.5}      \\ 
\bottomrule[0.75pt]
\end{tabular}}
\end{center}
\end{table}

\section{Additional Experiments in Varied Settings} \label{sec:addit-exp}

\begin{table*}[htbp]
\centering
\caption{Attack success rates (\%), Time consumption (min), and Memory usage (GB) of Ens vs. SEA with nine different baselines.}
\label{tab:classi}
\vskip 0.2in
\fontsize{7}{7}\selectfont
\setlength{\tabcolsep}{1.4mm}
\resizebox{0.96\linewidth}{!}{
\begin{tabular}{cccccccccccccc}
\toprule[0.75pt]
\multirow{2}{*}{Baseline}& \multirow{2}{*}{Attack} & \multicolumn{2}{c}{CNNs} & \multicolumn{2}{c}{ViTs} & \multicolumn{4}{c}{Defense} & \multirow{2}{*}{Avg.} & \multirow{2}{*}{\begin{tabular}[c]{@{}c@{}}Time \\(min)\end{tabular}} & \multirow{2}{*}{\begin{tabular}[c]{@{}c@{}}Memory \\(GB)\end{tabular}} \\ \cmidrule(lr){3-4} \cmidrule(lr){5-6} \cmidrule(lr){7-10} 
 &   & V19 & D121 & ViT-B & Swin-B & AT$_{ens4}$ & AT$_{ens}$ & HGD & RS  \\ 
\toprule[0.75pt]
                           & \textcolor{gray}{Ens(4,4)}                & \textcolor{gray}{78.9} & \textcolor{gray}{77.8} & \textcolor{gray}{29.9} & \textcolor{gray}{33.9} & \textcolor{gray}{41.9} & \textcolor{gray}{27.7} & \textcolor{gray}{34.3} & \textcolor{gray}{36.9} & \textcolor{gray}{45.2} & \textcolor{gray}{5.4}                                                  & \textcolor{gray}{18.4}                                                  \\
                           & Ens(4,3)                       & 75.0                        & 71.5                        & 27.1                        & 28.5                        & 37.6                        & 23.5                        & 27.0                        & 34.9                        & 40.6                        & 4.5                                                                         & 13.3                                                                         \\
                           & SEA(4,3)                   & \textbf{79.1}               & \textbf{78.3}               & \textbf{29.9}               & \textbf{33.8}               & \textbf{41.9}               & \textbf{27.6}               & \textbf{34.2}               & \textbf{36.7}               & \textbf{45.2}               & 4.6                                                                         & 13.3                                                                         \\
\multirow{-4}{*}{MI}       & SEA(4,2)                   & 78.9                        & 77.7                        & 29.7                        & 32.7                        & 41.9                        & 27.7                        & 31.0                        & 35.9                        & 44.4                        & \textbf{3.1}                                                                & \textbf{11.2}                                                                \\ \midrule
                           & \textcolor{gray}{Ens(4,4)}                & \textcolor{gray}{84.1} & \textcolor{gray}{81.1} & \textcolor{gray}{32.4} & \textcolor{gray}{34.7} & \textcolor{gray}{41.0} & \textcolor{gray}{26.6} & \textcolor{gray}{29.9} & \textcolor{gray}{37.9} & \textcolor{gray}{46.0} & \textcolor{gray}{5.4}                                                  & \textcolor{gray}{14.7}                                                  \\
                           & Ens(4,3)                       & 79.3                        & 76.9                        & 28.4                        & 28.3                        & 35.6                        & 22.0                        & 25.1                        & 36.6                        & 41.5                        & 4.5                                                                         & 12.4                                                                         \\
                           & SEA(4,3)                   & \textbf{85.0}               & \textbf{82.1}               & \textbf{31.7}               & \textbf{33.3}               & \textbf{39.8}               & \textbf{25.2}               & \textbf{27.3}               & \textbf{37.7}               & \textbf{45.3}               & 4.6                                                                         & 13.4                                                                         \\
\multirow{-4}{*}{NI}       & SEA(4,2)                   & 84.9                        & 81.7                        & 31.4                        & 31.3                        & 36.2                        & 24.7                        & 23.9                        & 36.1                        & 43.8                        & \textbf{3.1}                                                                & \textbf{11.3}                                                                \\ \midrule
                           & \textcolor{gray}{Ens(4,4)}                & \textcolor{gray}{88.4} & \textcolor{gray}{88.9} & \textcolor{gray}{49.6} & \textcolor{gray}{54.9} & \textcolor{gray}{67.9} & \textcolor{gray}{56.0} & \textcolor{gray}{61.7} & \textcolor{gray}{51.0} & \textcolor{gray}{64.8} & \textcolor{gray}{113.7}                                                & \textcolor{gray}{14.3}                                                  \\
                           & Ens(4,3)                       & 86.3                        & 86.0                        & 44.3                        & 49.6                        & 63.0                        & 50.8                        & 55.8                        & 46.5                        & 60.3                        & 85.9                                                                        & 12.6                                                                         \\
                           & SEA(4,3)                   & \textbf{87.8}               & \textbf{88.1}               & \textbf{48.0}               & \textbf{52.1}               & \textbf{65.3}               & \textbf{54.7}               & \textbf{59.7}               & \textbf{49.7}               & \textbf{63.2}               & 85.3                                                                        & 13.3                                                                         \\
\multirow{-4}{*}{VT}       & SEA(4,2)                   & 86.2                        & 85.8                        & 45.3                        & 50.5                        & 62.8                        & 48.6                        & 57.7                        & 46.1                        & 60.4                        & \textbf{57.0}                                                               & \textbf{11.6}                                                                \\ \midrule
                           & \textcolor{gray}{Ens(4,4)}                & \textcolor{gray}{95.6} & \textcolor{gray}{95.0} & \textcolor{gray}{73.6} & \textcolor{gray}{73.3} & \textcolor{gray}{88.1} & \textcolor{gray}{82.7} & \textcolor{gray}{84.1} & \textcolor{gray}{79.8} & \textcolor{gray}{84.0} & \textcolor{gray}{193.9}                                                & \textcolor{gray}{23.3}                                                  \\
                           & Ens(4,3)                       & 93.6                        & 93.9                        & 69.0                        & 67.3                        & 84.4                        & 78.8                        & 79.3                        & 75.6                        & 80.2                        & 142.4                                                                       & 21.0                                                                         \\
                           & SEA(4,3)                   & \textbf{94.6}               & \textbf{94.5}               & \textbf{72.4}               & \textbf{73.5}               & \textbf{87.1}               & \textbf{82.0}               & \textbf{82.1}               & \textbf{78.4}               & \textbf{83.1}               & 144.7                                                                       & 21.5                                                                         \\
\multirow{-4}{*}{PGN}      & SEA(4,2)                   & 94.0                        & 94.1                        & 68.7                        & 70.3                        & 84.5                        & 79.3                        & 78.7                        & 76.4                        & 80.8                        & \textbf{98.5}                                                               & \textbf{19.7}                                                                \\ \midrule
                           & \textcolor{gray}{Ens(4,4)}                & \textcolor{gray}{90.3} & \textcolor{gray}{91.5} & \textcolor{gray}{50.5} & \textcolor{gray}{53.2} & \textcolor{gray}{66.6} & \textcolor{gray}{52.3} & \textcolor{gray}{63.7} & \textcolor{gray}{50.6} & \textcolor{gray}{64.8} & \textcolor{gray}{5.5}                                                  & \textcolor{gray}{14.3}                                                  \\
                           & Ens(4,3)                       & 88.0                        & 89.1                        & 44.5                        & 42.9                        & 61.0                        & 42.0                        & 52.5                        & 47.4                        & 58.4                        & 4.1                                                                         & 12.0                                                                         \\
                           & SEA(4,3)                   & \textbf{90.0}               & \textbf{90.4}               & \textbf{48.3}               & \textbf{48.6}               & \textbf{64.9}               & \textbf{50.3}               & \textbf{61.4}               & \textbf{49.5}               & \textbf{62.9}               & 4.6                                                                         & 13.3                                                                         \\
\multirow{-4}{*}{DI}       & SEA(4,2)                   & 88.4                        & 88.4                        & 45.1                        & 42.6                        & 60.0                        & 44.0                        & 55.4                        & 45.7                        & 58.7                        & \textbf{3.1}                                                                & \textbf{11.4}                                                                \\ \midrule
                           & \textcolor{gray}{Ens(4,4)}                & \textcolor{gray}{82.7} & \textcolor{gray}{84.1} & \textcolor{gray}{41.0} & \textcolor{gray}{36.9} & \textcolor{gray}{59.4} & \textcolor{gray}{49.6} & \textcolor{gray}{62.8} & \textcolor{gray}{46.7} & \textcolor{gray}{57.9} & \textcolor{gray}{5.5}                                                  & \textcolor{gray}{14.3}                                                  \\
                           & Ens(4,3)                       & 77.9                        & 79.3                        & 37.4                        & 28.9                        & 59.1                        & 48.2                        & 54.3                        & 43.7                        & 53.6                        & 4.4                                                                         & 12.5                                                                         \\
                           & SEA(4,3)                   & \textbf{82.7}               & \textbf{82.6}               & \textbf{40.9}               & \textbf{35.9}               & \textbf{59.6}               & \textbf{49.5}               & \textbf{62.5}               & \textbf{46.6}               & \textbf{57.5}               & 4.6                                                                         & 13.2                                                                         \\
\multirow{-4}{*}{TI}       & SEA(4,2)                   & 82.1                        & 82.5                        & 40.8                        & 35.8                        & 58.2                        & 49.4                        & 62.0                        & 46.5                        & 57.2                        & \textbf{3.2}                                                                & \textbf{11.5}                                                                \\ \midrule
                           & \textcolor{gray}{Ens(4,4)}                & \textcolor{gray}{94.5} & \textcolor{gray}{95.9} & \textcolor{gray}{53.3} & \textcolor{gray}{53.8} & \textcolor{gray}{75.7} & \textcolor{gray}{58.9} & \textcolor{gray}{66.1} & \textcolor{gray}{57.3} & \textcolor{gray}{69.4} & \textcolor{gray}{27.3}                                                 & \textcolor{gray}{14.4}                                                  \\
                           & Ens(4,3)                       & 91.8                        & 93.3                        & 49.1                        & 47.4                        & 69.3                        & 52.3                        & 58.2                        & 53.8                        & 64.4                        & 20.6                                                                        & 12.6                                                                         \\
                           & SEA(4,3)                   & \textbf{94.0}               & \textbf{95.9}               & \textbf{52.5}               & \textbf{51.9}               & \textbf{75.1}               & \textbf{56.6}               & \textbf{64.6}               & \textbf{56.2}               & \textbf{68.4}               & 20.2                                                                        & 12.6                                                                         \\
\multirow{-4}{*}{SI}       & SEA(4,2)                   & 94.1                        & 95.8                        & 50.8                        & 50.3                        & 71.6                        & 55.6                        & 59.9                        & 54.6                        & 66.6                        & \textbf{13.9}                                                               & \textbf{11.4}                                                                \\ \midrule
                           & \textcolor{gray}{Ens(4,4)}                & \textcolor{gray}{95.8} & \textcolor{gray}{96.3} & \textcolor{gray}{71.5} & \textcolor{gray}{72.2} & \textcolor{gray}{85.8} & \textcolor{gray}{77.1} & \textcolor{gray}{81.1} & \textcolor{gray}{72.5} & \textcolor{gray}{81.5} & \textcolor{gray}{113.0}                                                & \textcolor{gray}{15.8}                                                  \\
                           & Ens(4,3)                       & 94.8                        & 94.9                        & 65.7                        & 65.4                        & 81.4                        & 72.6                        & 76.3                        & 68.7                        & 77.5                        & 86.0                                                                        & 13.6                                                                         \\
                           & SEA(4,3)                   & \textbf{95.7}               & \textbf{95.1}               & \textbf{69.8}               & \textbf{70.2}               & \textbf{84.3}               & \textbf{74.4}               & \textbf{78.8}               & \textbf{71.5}               & \textbf{80.0}               & 84.5                                                                        & 13.9                                                                         \\
\multirow{-4}{*}{SSA}      & SEA(4,2)                   & 94.8                        & 94.1                        & 66.9                        & 66.8                        & 81.7                        & 72.0                        & 74.1                        & 67.5                        & 77.2                        & \textbf{57.9}                                                               & \textbf{11.5}                                                                \\  \midrule
                       & \textcolor{gray}{Ens(4,4)} & \textcolor{gray}{91.9} & \textcolor{gray}{89.2} & \textcolor{gray}{41.2} & \textcolor{gray}{42.5} & \textcolor{gray}{52.3} & \textcolor{gray}{34.5} & \textcolor{gray}{38.1} & \textcolor{gray}{37.0} & \textcolor{gray}{53.3} & \textcolor{gray}{7.2} & \textcolor{gray}{22.8} \\
                        & Ens(4,3)        & 87.6                        & 86.1                        & 34.4                        & 32.6                        & 45.5                        & 28.6                        & 31.9                        & 34.5                        & 47.7                        & 4.8                        & 21.5                        \\
                        & SEA(4,3)    & \textbf{88.5}               & \textbf{87.2}               & \textbf{38.1}               & \textbf{38.0}               & \textbf{47.7}               & \textbf{33.2}               & \textbf{34.1}               & \textbf{37.0}               & \textbf{50.5}               & 5.0                        & 21.6                        \\
\multirow{-4}{*}{ghost} & SEA(4,2)    & 84.6                        & 84.1                        & 34.1                        & 31.5                        & 42.7                        & 26.0                        & 25.0                        & 33.9                        & 45.2                        & \textbf{3.6}               & \textbf{20.3}              \\
\bottomrule[0.75pt]
\end{tabular}}
\end{table*}

\begin{figure}[htbp]
  \centering
    \includegraphics[width=0.95\linewidth]{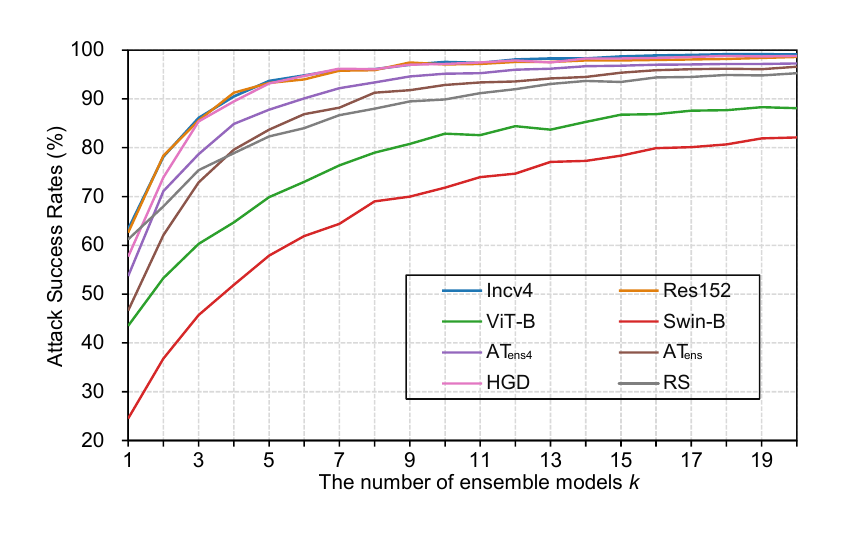}
    \caption{Attack success rates (\%) of transferring from an ensemble of $k$ models to eight different target models. TI is the transfer baseline.}
    \label{fig:moti-TI}
\end{figure}

\begin{table*}[!t]
\begin{center}
\caption{Transferability and resource efficiency of Ens vs. SEA with similar or diverse surrogates. TI is the transfer baseline.}
\label{tab:diver-TI}
\fontsize{8}{8.5}\selectfont
\setlength{\tabcolsep}{1.4mm}
\resizebox{0.96\linewidth}{!}{
\begin{tabular}{cccccccccccccc}
\toprule[0.8pt]
\multirow{2}{*}{Surrogates}& \multirow{2}{*}{Attack} & \multicolumn{2}{c}{CNNs} & \multicolumn{2}{c}{ViTs} &  \multicolumn{4}{c}{Defense} & \multirow{2}{*}{Avg.} & \multirow{2}{*}{\begin{tabular}[c]{@{}c@{}}Time \\(min)\end{tabular}} & \multirow{2}{*}{\begin{tabular}[c]{@{}c@{}}Memory \\(GB)\end{tabular}} \\ \cmidrule(lr){3-4} \cmidrule(lr){5-6} \cmidrule(lr){7-10} 
 &  & Incv4 & Res152 & ViT-B & Swin-B & AT$_{ens4} $ & AT$_{ens} $ & HGD & RS  \\ 
\toprule[0.8pt]
                           & \textcolor{gray}{Ens(4,4)} & \textcolor{gray}{88.0} & \textcolor{gray}{90.8} & \textcolor{gray}{61.8} & \textcolor{gray}{48.7} & \textcolor{gray}{84.0} & \textcolor{gray}{76.7} & \textcolor{gray}{90.4} & \textcolor{gray}{82.4} & \textcolor{gray}{77.9} & \textcolor{gray}{1.9} & \textcolor{gray}{7.2}  \\
                           & Ens(4,3)        & 81.7                        & 86.7                        & 57.6                        & 40.1                        & 78.2                        & 70.9                        & 85.9                        & 78.6                        & 72.5                        & 1.6                        & 6.8                         \\
                           & SEA(4,3)    & \textbf{88.0}               & \textbf{91.1}               & \textbf{61.1}               & \textbf{46.8}               & \textbf{83.2}               & \textbf{76.6}               & \textbf{90.9}               & \textbf{82.3}               & \textbf{77.5}               & 1.5                        & 6.8                         \\
\multirow{-4}{*}{resnet}   & SEA(4,2)    & 86.5                        & 89.5               & 59.7                        & 43.7                        & 81.4                        & 74.0                        & 88.7                        & \textbf{82.3}               & 75.7                        & \textbf{1.2}               & \textbf{6.4}                \\ \midrule
                           & \textcolor{gray}{Ens(4,4)} & \textcolor{gray}{89.3} & \textcolor{gray}{77.9} & \textcolor{gray}{52.9} & \textcolor{gray}{43.5} & \textcolor{gray}{70.8} & \textcolor{gray}{62.5} & \textcolor{gray}{78.7} & \textcolor{gray}{72.5} & \textcolor{gray}{68.5} & \textcolor{gray}{2.7} & \textcolor{gray}{10.6} \\
                           & Ens(4,3)        & 86.7                        & 74.6                        & 49.4                        & 38.7                        & 67.8                        & 58.0                        & 72.2                        & 68.8                        & 64.5                        & 2.1                        & 10.1                        \\
                           & SEA(4,3)    & \textbf{89.1}               & \textbf{77.0}               & \textbf{53.0}               & \textbf{42.4}               & \textbf{70.4}               & \textbf{60.7}               & \textbf{76.5}               & \textbf{71.8}               & \textbf{67.6}               & 2.1                        & 8.9                         \\
\multirow{-4}{*}{vgg}      & SEA(4,2)    & 87.2               & 76.8                        & 52.3                        & 40.6                        & 69.2                        & 58.7                        & 75.0                        & 71.7                        & 66.4                        & \textbf{1.1}               & \textbf{7.5}                \\ \midrule
                           & \textcolor{gray}{Ens(4,4)} & \textcolor{gray}{90.7} & \textcolor{gray}{92.2} & \textcolor{gray}{66.0} & \textcolor{gray}{53.5} & \textcolor{gray}{85.1} & \textcolor{gray}{81.9} & \textcolor{gray}{92.5} & \textcolor{gray}{81.4} & \textcolor{gray}{80.4} & \textcolor{gray}{3.2} & \textcolor{gray}{16.3} \\
                           & Ens(4,3)        & 86.9                        & 89.1               & 62.6                        & 47.8                        & 81.8                        & 75.3                        & 89.8                        & 77.8                        & 76.4                        & 2.5                        & 13.7                        \\
                           & SEA(4,3)    & \textbf{90.6}               & \textbf{91.8}               & \textbf{64.7}               & \textbf{52.3}               & \textbf{84.8}               & \textbf{80.8}               & \textbf{91.8}               & \textbf{81.6}               & \textbf{79.8}               & 2.5                        & 13.6                        \\
\multirow{-4}{*}{densenet} & SEA(4,2)    & 89.5                        & 91.1               & 63.5                        & 48.2                        & 83.6                        & 79.0                        & 90.4                        & 81.1                        & 78.3                       & \textbf{1.9}               & \textbf{10.7}               \\ \midrule
                           & \textcolor{gray}{Ens(4,4)} & \textcolor{gray}{88.8} & \textcolor{gray}{90.0} & \textcolor{gray}{64.0} & \textcolor{gray}{52.7} & \textcolor{gray}{85.3} & \textcolor{gray}{85.2} & \textcolor{gray}{90.9} & \textcolor{gray}{74.7} & \textcolor{gray}{79.0} & \textcolor{gray}{6.6} & \textcolor{gray}{22.0} \\
                           & Ens(4,3)        & 83.2                        & 86.9                        & 58.1                        & 46.4                        & 80.1                        & 77.1                        & 86.0                        & 72.1                        & 73.7                        & 5.2                        & 19.9                        \\
                           & SEA(4,3)    & \textbf{88.4}               & \textbf{89.9}               & \textbf{63.5}               & \textbf{51.6}               & \textbf{85.8}               & \textbf{83.6}               & \textbf{90.2}               & \textbf{74.9}               & \textbf{78.5}               & 5.1                        & 20.1                        \\
\multirow{-4}{*}{dpn}      & SEA(4,2)    & 87.8                       & 88.8                       & 61.4                        & 51.3                        & 84.1                        & 81.6                        & 88.4                        & 74.8                        & 77.3                        & \textbf{3.7}               & \textbf{15.4}               \\ \midrule
                           & \textcolor{gray}{Ens(4,4)} & \textcolor{gray}{91.2} & \textcolor{gray}{92.5} & \textcolor{gray}{68.2} & \textcolor{gray}{57.9} & \textcolor{gray}{87.2} & \textcolor{gray}{83.5} & \textcolor{gray}{92.8} & \textcolor{gray}{82.5} & \textcolor{gray}{82.0} & \textcolor{gray}{4.5} & \textcolor{gray}{17.8} \\
                           & Ens(4,3)        & 88.3                        & 89.8                       & 60.8                        & 49.1                        & 81.6                        & 77.8                        & 90.2                        & 78.3                        & 77.0                        & 3.6                        & 16.5                        \\
                           & SEA(4,3)    & \textbf{91.7}               & \textbf{91.8}               & \textbf{66.3}               & \textbf{56.4}               & \textbf{86.5}               & \textbf{83.1}               & \textbf{92.2}               & \textbf{82.1}               & \textbf{81.3}               & 3.5                        & 15.9                        \\
\multirow{-4}{*}{diverse}  & SEA(4,2)    & 91.5                        & 91.3                       & 63.7                        & 54.7                        & 85.4                        & 81.9                        & 91.0                        & 80.8                        & 80.0                        & \textbf{2.6}               & \textbf{14.1}               \\
\bottomrule[0.8pt]
\end{tabular}}
\end{center}
\end{table*}

\begin{table*}[!t]
\centering
\caption{Transferability and resource efficiency of Ens vs. SEA integrated with four advanced ensemble strategies.
}
\label{tab:ensem}
\vskip 0.1in
\fontsize{8}{8.5}\selectfont
\setlength{\tabcolsep}{1.3mm}
\resizebox{\linewidth}{!}{
\begin{tabular}{ccccccccccccccc}
\toprule[0.85pt]
\multirow{2}{*}{Base}& \multirow{2}{*}{\begin{tabular}[c]{@{}c@{}}Ensemble \\strategy\end{tabular}} & \multirow{2}{*}{Attack} & \multicolumn{2}{c}{CNNs} & \multicolumn{2}{c}{ViTs} & \multicolumn{4}{c}{Defense} & \multirow{2}{*}{Avg.} & \multirow{2}{*}{\begin{tabular}[c]{@{}c@{}}Time \\(min)\end{tabular}} & \multirow{2}{*}{\begin{tabular}[c]{@{}c@{}}Memory \\(GB)\end{tabular}} \\ \cmidrule(lr){4-5} \cmidrule(lr){6-7} \cmidrule(lr){8-11} 
 &  &   & V19 & D121 & ViT-B & Swin-B & AT3$_{ens4} $ & AT$_{ens} $ & HGD & RS  \\ 
\toprule[0.85pt]
\multirow{16}{*}{MI}&  & \textcolor{gray}{Ens(4,4)} & \textcolor{gray}{91.3} & \textcolor{gray}{91.2} & \textcolor{gray}{41.3} & \textcolor{gray}{47.9} & \textcolor{gray}{55.5} & \textcolor{gray}{37.5} & \textcolor{gray}{43.8} & \textcolor{gray}{42.6} & \textcolor{gray}{56.4} & \textcolor{gray}{46.4}  & \textcolor{gray}{15.2} \\
                      &  & Ens(4,3)        & 87.0                        & 86.1                        & 34.9                        & 38.4                        & 45.2                        & 30.7                        & 35.4                        & 39.1                        & 49.6                        & 33.9                         & 14.8                        \\
                      &  & SEA(4,3)    & \textbf{89.3}               & \textbf{88.7}               & \textbf{38.8}               & \textbf{44.1}               & \textbf{51.1}               & \textbf{33.3}               & \textbf{41.8}               & \textbf{42.1}               & \textbf{53.7}               & 34.6                         & 14.4                        \\
&\multirow{-4}{*}{SVRE}  & SEA(4,2)    & 85.7                        & 84.9                        & 34.0                        & 37.5                        & 45.3                        & 28.6                        & 34.5                        & 37.2                        & 48.5                        & \textbf{23.2}                & \textbf{14.1}               \\ \cmidrule{2-14}
                     &   & \textcolor{gray}{Ens(4,4)} & \textcolor{gray}{79.0} & \textcolor{gray}{79.1} & \textcolor{gray}{30.8} & \textcolor{gray}{33.4} & \textcolor{gray}{42.7} & \textcolor{gray}{29.1} & \textcolor{gray}{35.6} & \textcolor{gray}{36.8} & \textcolor{gray}{45.8} & \textcolor{gray}{53.2}  & \textcolor{gray}{19.2} \\
                    &    & Ens(4,3)        & 74.8                        & 71.4                        & 26.2                        & 27.1                        & 32.6                        & 20.7                        & 21.9                        & 32.0                        & 38.3                        & 37.6                         & 14.2                        \\
                     &   & SEA(4,3)    & \textbf{78.4}               & \textbf{76.8}               & \textbf{31.2}               & \textbf{32.4}               & \textbf{37.0}               & \textbf{24.9}               & \textbf{27.4}               & \textbf{35.7}               & \textbf{43.0}               & 33.9                         & 14.1                        \\
&\multirow{-4}{*}{AdaEA} & SEA(4,2)    & 75.7                        & 74.0                        & 28.7                        & 31.9                        & 32.4                        & 21.0                        & 20.0                        & 34.3                        & 39.8                        & \textbf{24.5}                & \textbf{9.5}                \\ \cmidrule{2-14}
                    &    & \textcolor{gray}{Ens(4,4)} & \textcolor{gray}{91.7} & \textcolor{gray}{92.4} & \textcolor{gray}{41.7} & \textcolor{gray}{48.1} & \textcolor{gray}{53.9} & \textcolor{gray}{36.2} & \textcolor{gray}{40.8} & \textcolor{gray}{43.2} & \textcolor{gray}{56.0} & \textcolor{gray}{100.3} & \textcolor{gray}{15.2} \\
                    &    & Ens(4,3)        & 89.2                        & 86.9                        & 37.0                        & 37.9                        & 44.6                        & 28.5                        & 31.7                        & 38.7                        & 49.3                        & 67.1                         & 13.9                        \\
                    &    & SEA(4,3)    & \textbf{90.0}               & \textbf{88.3}               & \textbf{37.8}               & \textbf{39.5}               & \textbf{47.5}               & \textbf{31.0}               & \textbf{34.9}               & \textbf{39.4}               & \textbf{51.1}               & 65.9                         & 13.3                        \\
&\multirow{-4}{*}{SMER}  & SEA(4,2)    & 83.9                        & 82.8                        & 32.5                        & 31.9                        & 41.8                        & 25.4                        & 29.9                        & 37.7                        & 45.7                        & \textbf{33.7}                & \textbf{12.3}               \\ \cmidrule{2-14}
                    &    & \textcolor{gray}{Ens(4,4)} & \textcolor{gray}{92.9} & \textcolor{gray}{91.1} & \textcolor{gray}{33.5} & \textcolor{gray}{33.3} & \textcolor{gray}{32.0} & \textcolor{gray}{19.2} & \textcolor{gray}{27.1} & \textcolor{gray}{43.0} & \textcolor{gray}{46.5} & \textcolor{gray}{11.5}  & \textcolor{gray}{14.5} \\
                    &    & Ens(4,3)        & 85.5                        & 80.1                        & 26.5                        & 23.3                        & 26.8                        & 16.3                        & 25.2                        & 37.9                        & 40.2                        & 8.7                          & 13.5                        \\
                    &    & SEA(4,3)    & \textbf{91.2}               & \textbf{88.3}               & \textbf{33.3}               & \textbf{32.5}               & \textbf{33.3}               & \textbf{20.0}               & \textbf{27.6}               & \textbf{43.1}               & \textbf{46.2}               & 8.7                          & 13.3                        \\
&\multirow{-4}{*}{CWA}   & SEA(4,2)    & 88.0                        & 86.5                        & 29.8                        & 28.2                        & 30.5                        & 18.6                        & 27.5                        & 41.3                        & 43.8                        & \textbf{6.2}                 & \textbf{11.5}               \\ \cmidrule{1-14}
\multirow{16}{*}{TI} &    & \textcolor{gray}{Ens(4,4)} & \textcolor{gray}{92.0} & \textcolor{gray}{93.2} & \textcolor{gray}{53.7} & \textcolor{gray}{45.1} & \textcolor{gray}{74.0} & \textcolor{gray}{66.3} & \textcolor{gray}{78.7} & \textcolor{gray}{56.3} & \textcolor{gray}{69.9} & \textcolor{gray}{44.7}  & \textcolor{gray}{15.4} \\
                    &    & Ens(4,3)        & 89.3                        & 90.2                        & 48.1                        & 39.4                        & 67.2                        & 58.6                        & 71.2                        & 51.3                        & 64.4                        & 34.2                         & 14.9                        \\
                    &    & SEA(4,3)    & \textbf{91.1}               & \textbf{92.3}               & \textbf{53.1}               & \textbf{44.8}               & \textbf{73.2}               & \textbf{64.5}               & \textbf{78.5}               & \textbf{57.2}               & \textbf{69.3}               & 34.3                         & 14.5                        \\
& \multirow{-4}{*}{SVRE}  & SEA(4,2)    & 88.3                        & 91.0                        & 47.4                        & 40.2                        & 65.5                        & 55.9                        & 71.3                        & 51.7                        & 63.9                        & \textbf{23.4}                & \textbf{14.3}               \\ \cmidrule{2-14}
                    &    & \textcolor{gray}{Ens(4,4)} & \textcolor{gray}{83.2} & \textcolor{gray}{84.2} & \textcolor{gray}{42.1} & \textcolor{gray}{35.5} & \textcolor{gray}{60.0} & \textcolor{gray}{51.2} & \textcolor{gray}{63.6} & \textcolor{gray}{47.3} & \textcolor{gray}{58.4} & \textcolor{gray}{57.3}  & \textcolor{gray}{19.2} \\
                    &    & Ens(4,3)        & 78.1                        & 79.4                        & 37.1                        & 28.4                        & 52.1                        & 43.2                        & 52.7                        & 43.6                        & 51.8                        & 36.5                         & 14.2                        \\
                    &    & SEA(4,3)    & \textbf{83.4}               & \textbf{83.5}               & \textbf{40.9}               & \textbf{36.8}               & \textbf{59.2}               & \textbf{50.0}               & \textbf{60.2}               & \textbf{46.7}               & \textbf{57.6}               & 35.5                         & 14.1                        \\
&\multirow{-4}{*}{AdaEA} & SEA(4,2)    & 80.4                        & 80.9                        & 40.2                        & 30.6                        & 54.4                        & 46.9                        & 54.9                        & 43.7                        & 54.0                        & \textbf{23.8}                & \textbf{9.4}                \\ \cmidrule{2-14}
                    &    & \textcolor{gray}{Ens(4,4)} & \textcolor{gray}{93.1} & \textcolor{gray}{93.9} & \textcolor{gray}{54.7} & \textcolor{gray}{48.2} & \textcolor{gray}{76.7} & \textcolor{gray}{67.6} & \textcolor{gray}{78.9} & \textcolor{gray}{58.8} & \textcolor{gray}{71.5} & \textcolor{gray}{111.4} & \textcolor{gray}{15.2} \\
                    &    & Ens(4,3)        & 90.0                        & 90.3                        & 47.4                        & 39.1                        & 65.9                        & 57.2                        & 68.4                        & 51.2                        & 63.7                        & 67.4                         & 14.1                        \\
                    &    & SEA(4,3)    & \textbf{91.4}               & \textbf{92.5}               & \textbf{49.4}               & \textbf{42.9}               & \textbf{68.8}               & \textbf{60.9}               & \textbf{71.3}               & \textbf{53.9}               & \textbf{66.4}               & 67.4                         & 14.0                        \\
&\multirow{-4}{*}{SMER}  & SEA(4,2)    & 86.9                        & 87.7                        & 43.0                        & 35.0                        & 59.9                        & 49.7                        & 62.8                        & 47.6                        & 59.1                        & \textbf{34.2}                & \textbf{12.3}               \\ \cmidrule{2-14}
                    &    & \textcolor{gray}{Ens(4,4)} & \textcolor{gray}{89.5} & \textcolor{gray}{86.5} & \textcolor{gray}{33.2} & \textcolor{gray}{28.3} & \textcolor{gray}{39.0} & \textcolor{gray}{27.3} & \textcolor{gray}{26.2} & \textcolor{gray}{41.2} & \textcolor{gray}{46.4} & \textcolor{gray}{11.3}  & \textcolor{gray}{14.5} \\
                    &    & Ens(4,3)        & 83.0                        & 73.5                        & 26.7                        & 22.2                        & 29.7                        & 19.2                        & 17.4                        & 34.7                        & 38.3                        & 8.7                          & 13.5                        \\
                    &    & SEA(4,3)    & \textbf{89.1}               & \textbf{86.9}               & \textbf{31.9}               & \textbf{27.7}               & \textbf{38.2}               & \textbf{26.5}               & \textbf{26.6}               & \textbf{41.4}               & \textbf{46.0}               & 8.6                          & 13.3                        \\
&\multirow{-4}{*}{CWA}   & SEA(4,2)    & 85.6                        & 84.6                        & 31.8                        & 26.9                        & 38.1                        & 25.9                        & 26.5                        & \textbf{41.4}               & 45.1                        & \textbf{5.9}                 & \textbf{11.6}               \\ 
\bottomrule[0.85pt]

\end{tabular}}

\end{table*}

We provide additional experimental results to demonstrate the validity of the conclusions presented in the main text.
These experiments cover varied \textbf{number of surrogate models}, \textbf{surrogate model architectures}, and \textbf{ensemble strategies}. The setting of 4 surrogate models is consistent with that of previous works \cite{Lin2020Nesterov,Wang2021Variance}.
For experiments that have been reported with the MI baseline in the main text, we additionally evaluate TI, a popular input transformation-based transferable attack.
We also add another popular surrogate model refinement attack baseline~\cite{li2020learning,wu2020skip,zhao2025revisiting}, Ghost~\cite{li2020learning}.
Ghost randomly drops out neurons in the surrogate model.

The results of these experiments are illustrated in Figure \ref{fig:moti-TI}, Table \ref{tab:classi}, and Table \ref{tab:diver-TI}. 
As can be seen, all the conclusions drawn from these additional results are consistent with those in Figure~\ref{fig:moti}, Table~\ref{tab:20-4}, and Table~\ref{tab:diver}.

\section{SEA Integrated with Advanced Ensemble Strategies} 
\label{sec:ens-strategy}
\vskip -0.02in

In the main text, we have compared our SEA to advanced model ensemble strategies: SVRE, AdaEA, SMER, and CWA.
Here we further show that SEA can integrate them for further improvement of transferability. 
As can be seen from Table~\ref{tab:ensem}, our SEA consistently outperforms Ens by a large margin.
For example, SEA boosts the transferability of the state-of-the-art method, CWA, by 6.0\% on average.

\begin{figure*}[tbp]
    \centering
    \begin{minipage}{0.49\textwidth}
        \centering
        \includegraphics[width=\linewidth]{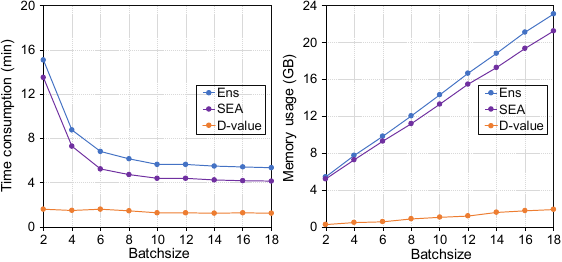}
        (a) MI
    \end{minipage}
    \begin{minipage}{0.49\textwidth}
        \centering
        \includegraphics[width=\linewidth]{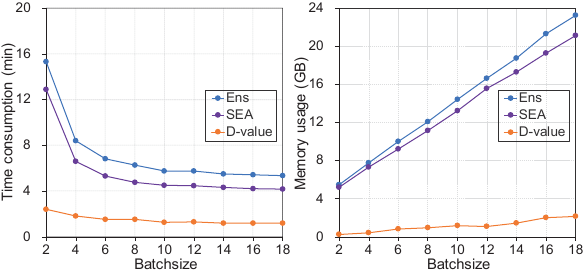}
        (b) TI
    \end{minipage}
    \caption{Time consumption and memory usage of Ens vs. SEA when varying the batch size. D-value represents the difference between Ens and SEA. MI and TI are the transfer baselines.}
    \label{fig:time-me}
\end{figure*}

\section{The Impact of Batchsize on Resource Efficiency}
\label{sec:batchsize}

The setting of optimization batch size may have an impact on the actual consumption of resources.
To investigate this impact, we conduct experiments with the upper bound Ens(4,4) and our SEA(4,3) under varied batch sizes.
As can be seen from Figure~\ref{fig:time-me}, as expected, our SEA(4,3) consistently reduces the costs of Ens(4,4).
Specifically, the improvement is more sensitive to the memory usage than the time consumption.
In addition, as the batch size increases, the difference in time consumption between these two attacks becomes smaller, but the difference in memory usage becomes larger.
This means that in practice, one should adjust the batch size based on the actual constraints in time consumption vs. memory usage to better leverage the benefit of our SEA.

\section{Ensemble of ViTs and CNNs}
\label{sec:integrating}

\begin{table*}[!t]
\begin{center}
\caption{Attack success rates (\%) of Ens vs. SEA with MI under 3CNNs vs. 3CNNs+1CNN/ViT model settings.} 
\label{tab:vit-cnn}
\vskip 0.1in
\begin{tabular}{c|c|cccc}
\toprule[0.9pt]
Attack & Surrogate Model & MobileV2 & Dpn131 & DeiT-B & PiT-B \\ \midrule[0.9pt]
Ens    & Res101/Dense161/Vgg11 & 96.1 & 94.7 & 58.8 & 66.3 \\ 
SEA    & +Res50              & \textbf{97.2} & \textbf{97.5} & 63.7 & 69.9 \\ 
SEA    & +Dense201           & 96.9 & 97.3 & 66.9 & 73.8 \\ 
SEA    & +ViT-B              & 97.1 & 95.9 & 84.4 & 77.9 \\
SEA    & +Cait-S             & 96.8 & 96.1 & \textbf{93.8} & \textbf{84.0} \\ 
\bottomrule[0.9pt]
\end{tabular}
\end{center}
\end{table*}

We further validate the effectiveness of SEA under a more challenging ensemble setting involving both CNNs and ViTs, as opposite effects may occur when highly diverse models are selected \cite{chen2023AdaEA}. Specifically, Ens ensembles three CNNs, whereas SEA adds a model with a different architecture to the ensemble. The results in Table \ref{tab:vit-cnn} show that even when the ensemble models have diverse architectures (CNN and ViT), SEA still outperforms Ens on all target models. More specifically, incorporating an additional CNN (ViT) into the ensemble enhances the attack performance against CNN (ViT) target models.

\section{Visualizations of Adversarial Images on APIs and VLMs}
\label{sec:visualization}

\begin{figure*}[htbp]
  \centering
    \includegraphics[width=0.9\linewidth]{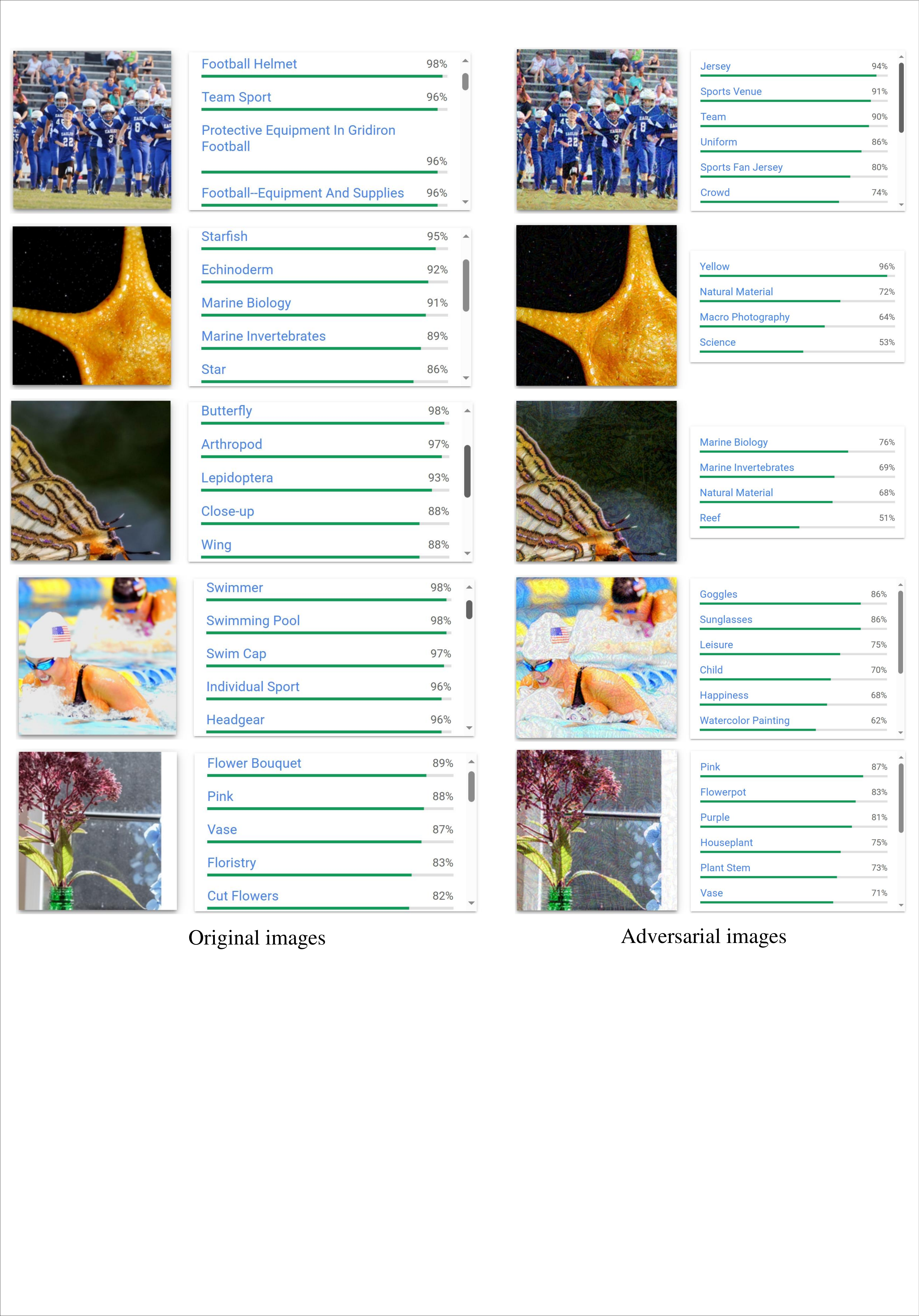}
    \caption{Original images and SEA adversarial images on Google Cloud Vision.}
    \label{fig:google-vision}
\end{figure*}

\begin{figure*}[htbp]
  \centering
    \includegraphics[width=0.95\linewidth]{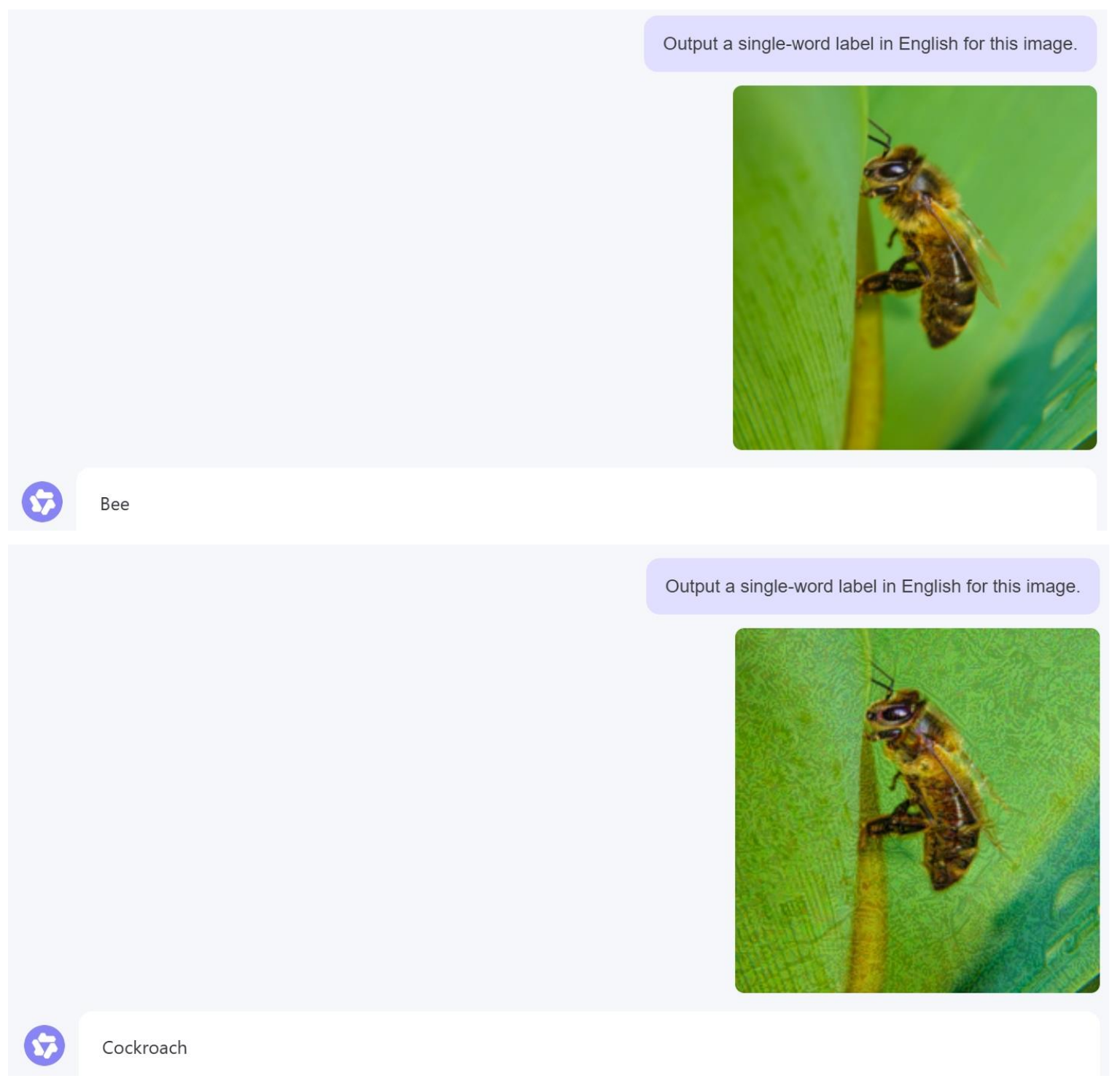}
    \caption{Original image and SEA adversarial image on Qwen with the prompt "Output a single-word label in English for this image".}
    \label{fig:Qwen-single}
\end{figure*}

\begin{figure*}[htbp]
  \centering
    \includegraphics[width=0.95\linewidth]{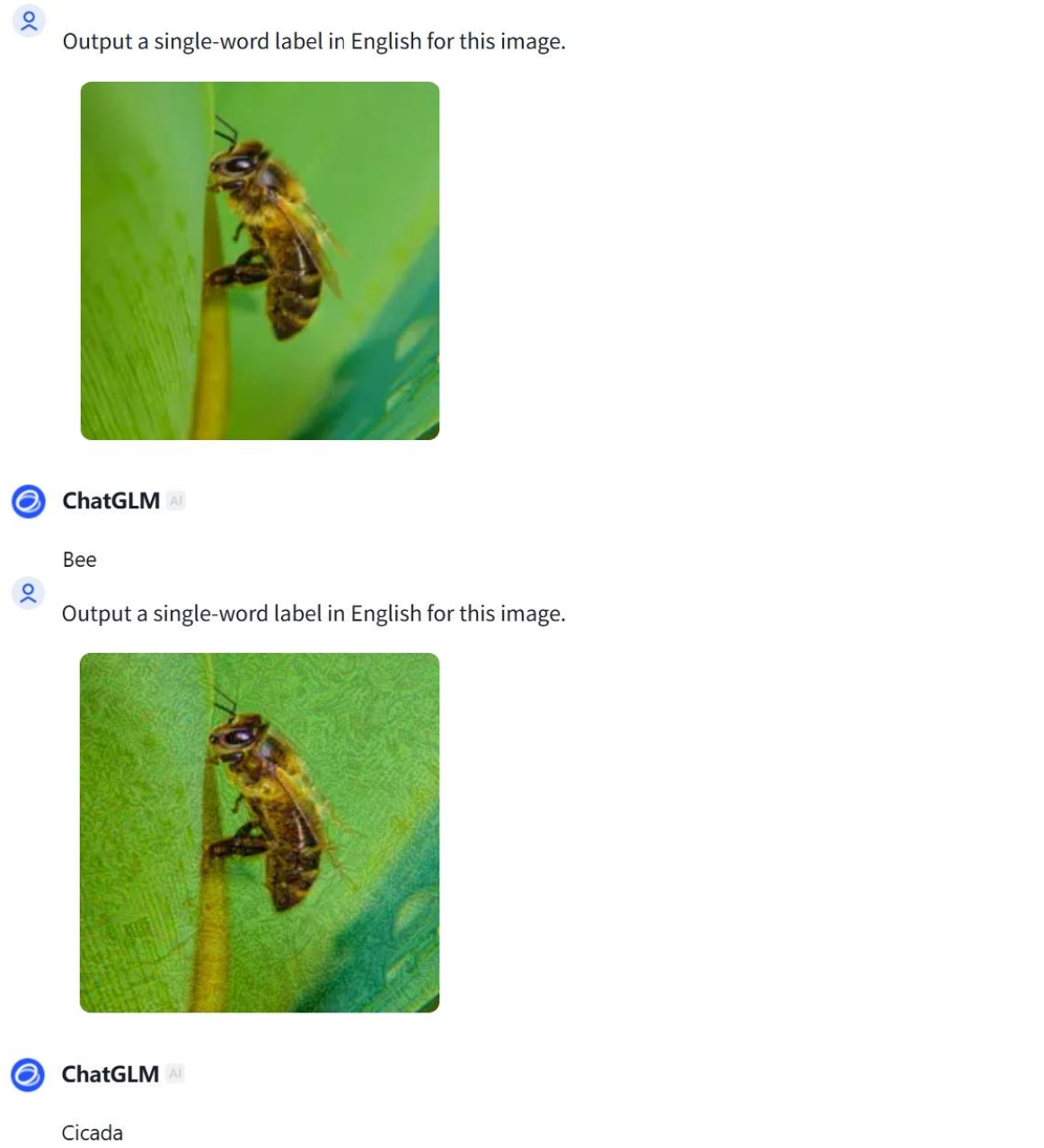}
    \caption{Original image and SEA adversarial image on ChatGLM with the prompt "Output a single-word label in English for this image".}
    \label{fig:ChatGLM-single}
\end{figure*}

\begin{figure*}[htbp]
  \centering
    \includegraphics[width=0.95\linewidth]{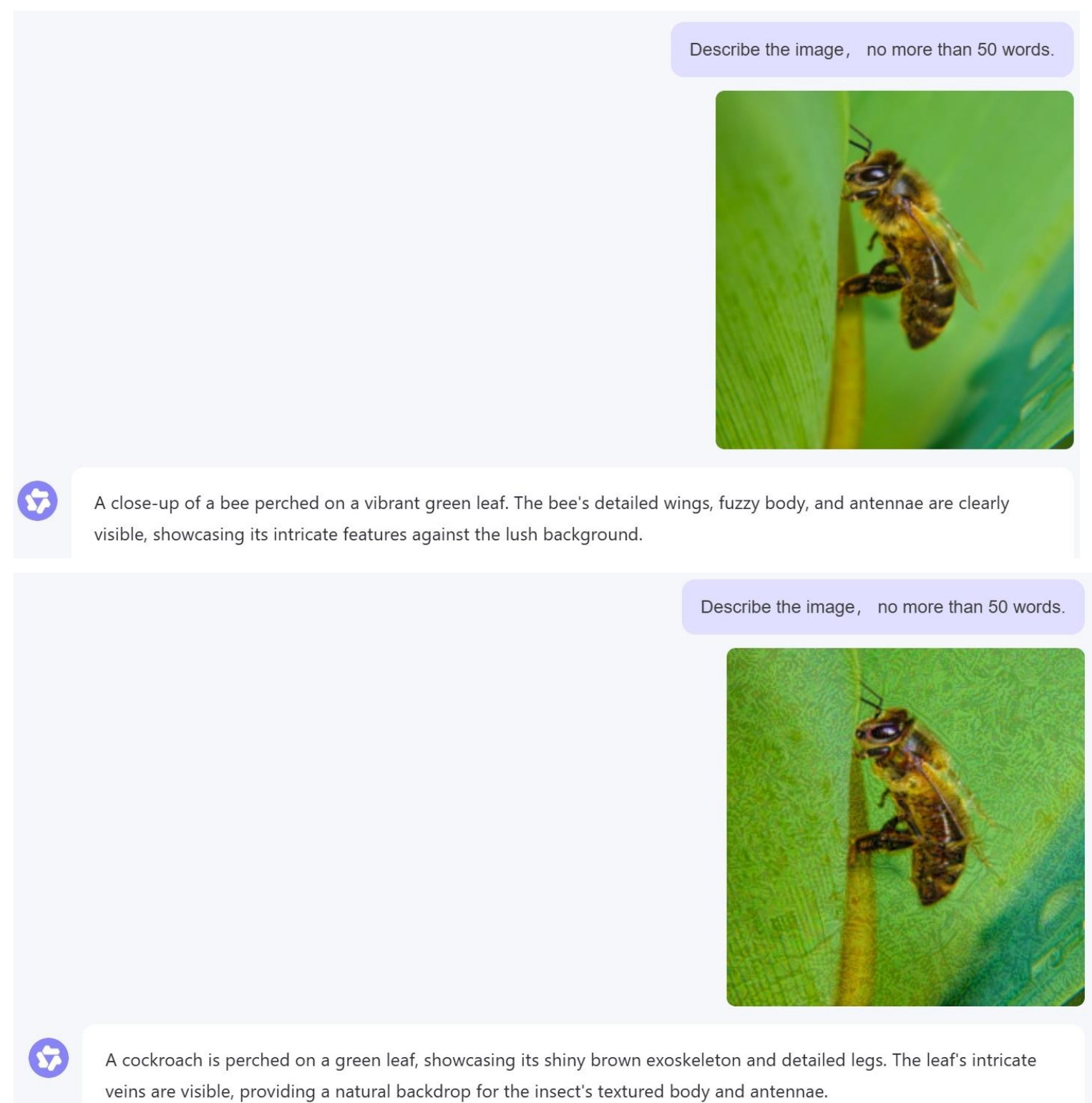}
    \caption{Original image and SEA adversarial image on Qwen with the prompt "Describe the image, no more than 50 words".}
    \label{fig:Qwen-describe}
\end{figure*}

\begin{figure*}[htbp]
  \centering
    \includegraphics[width=0.95\linewidth]{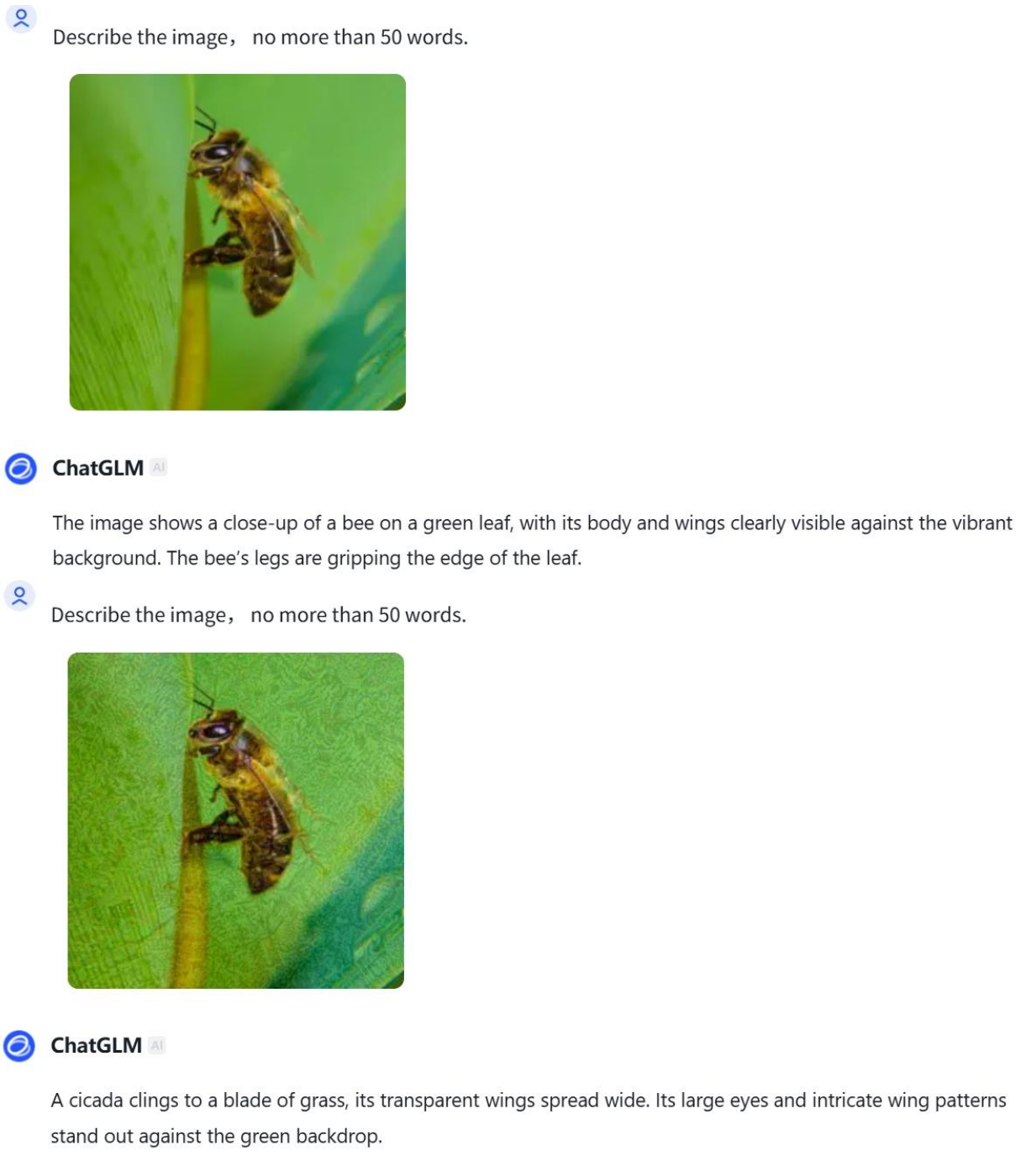}
    \caption{Original image and SEA adversarial image on ChatGLM with the prompt "Describe the image, no more than 50 words".}
    \label{fig:ChatGLM-describe}
\end{figure*}

In Section \ref{sec:Ens(20)}, the quantitative results on APIs and LVLMs (presented in Tables \ref{tab:api_LVLMs}) have demonstrated that SEA outperforms Ens. Here, we provide visualizations of the original images and the adversarial images crafted by SEA, as shown in Figures \ref{fig:google-vision}, \ref{fig:Qwen-single}, \ref{fig:ChatGLM-single}, \ref{fig:Qwen-describe}, and \ref{fig:ChatGLM-describe}. It can be observed that even ensembling small image classification models can effectively attack APIs and LVLMs, posing significant challenges in the development of robust vision APIs and LVLMs.

\end{document}